\documentclass[10pt,twocolumn,letterpaper]{article}

\usepackage{cvpr}              

\usepackage{graphicx}
\usepackage{amsmath}
\usepackage{amssymb}
\usepackage{booktabs}
\usepackage{bm}
\usepackage{arydshln}
\usepackage{siunitx}
\makeatletter
\def\adl@drawiv#1#2#3{%
        \hskip.5\tabcolsep
        \xleaders#3{#2.5\@tempdimb #1{1}#2.5\@tempdimb}%
                #2\z@ plus1fil minus1fil\relax
        \hskip.5\tabcolsep}
\newcommand{\cdashlinelr}[1]{%
  \noalign{\vskip\aboverulesep
           \global\let\@dashdrawstore\adl@draw
           \global\let\adl@draw\adl@drawiv}
  \cdashline{#1}
  \noalign{\global\let\adl@draw\@dashdrawstore
           \vskip\belowrulesep}}
\makeatother
\usepackage[pagebackref,breaklinks,colorlinks]{hyperref}

\usepackage[capitalize]{cleveref}
\crefname{section}{Sec.}{Secs.}
\Crefname{section}{Section}{Sections}
\Crefname{table}{Table}{Tables}
\crefname{table}{Tab.}{Tabs.}

\def\cvprPaperID{8412} 
\def\confName{CVPR}
\def\confYear{2023}

\begin{document}

\title{DyLiN: Making Light Field Networks Dynamic
}

\author{Heng Yu$^{1}$ \quad Joel Julin$^{1}$  \quad Zolt\'{a}n \'{A}. Milacski$^{1}$\quad Koichiro Niinuma$^{2}$ \quad L\'{a}szl\'{o} A. Jeni$^{1}$ \vspace{4pt}\\
	$^1$Robotics Institute, Carnegie Mellon University \quad
    $^2$Fujitsu Research of America \\
    {\tt\small \{hengyu, jjulin, zmilacsk\}@andrew.cmu.edu} \quad {\tt\small kniinuma@fujitsu.com} \quad {\tt\small laszlojeni@cmu.edu} \\
}



\twocolumn[{%
\renewcommand\twocolumn[1][]{#1}%
\maketitle
\vspace{-1.04cm}
\begin{center}
    \centering
    \captionsetup{type=figure}
    \begin{minipage}{0.15\textwidth}
    \centering
    \includegraphics[width=0.99\textwidth]{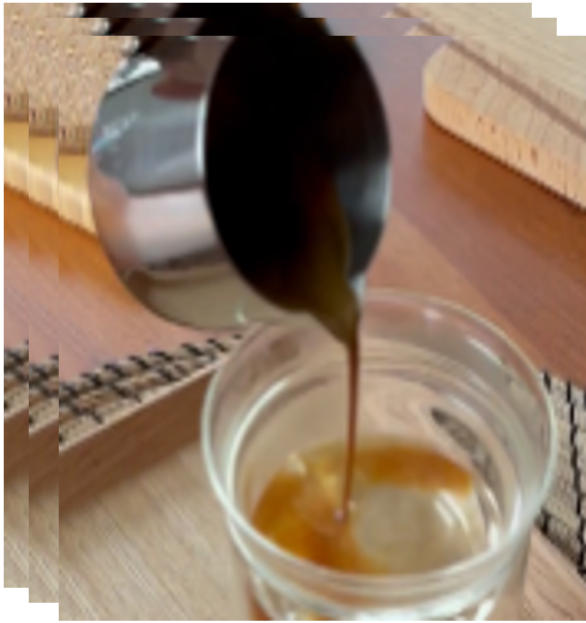} \\
    Ground Truth \\
    Video
    \end{minipage}
    \begin{minipage}{0.15\textwidth}
    \centering
    \includegraphics[width=0.99\textwidth]{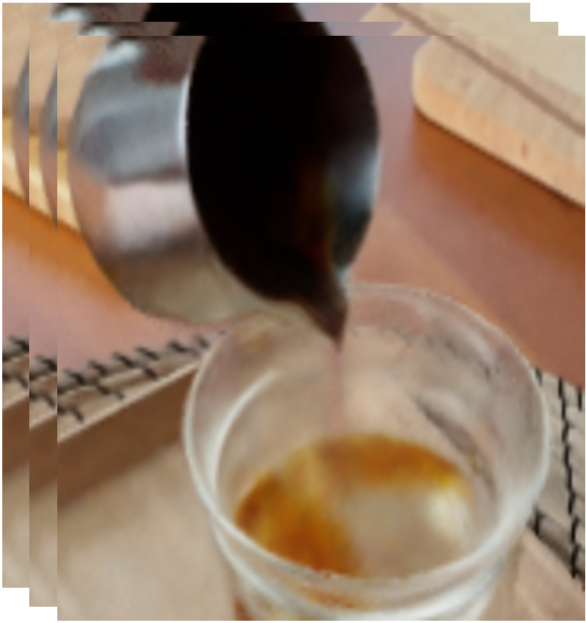} \\
    HyperNeRF \cite{park2021hypernerf}  \\
    Render time: $\SI{3}{\second}$
    \end{minipage}
    \begin{minipage}{0.15\textwidth}
    \centering
    \includegraphics[width=0.99\textwidth]{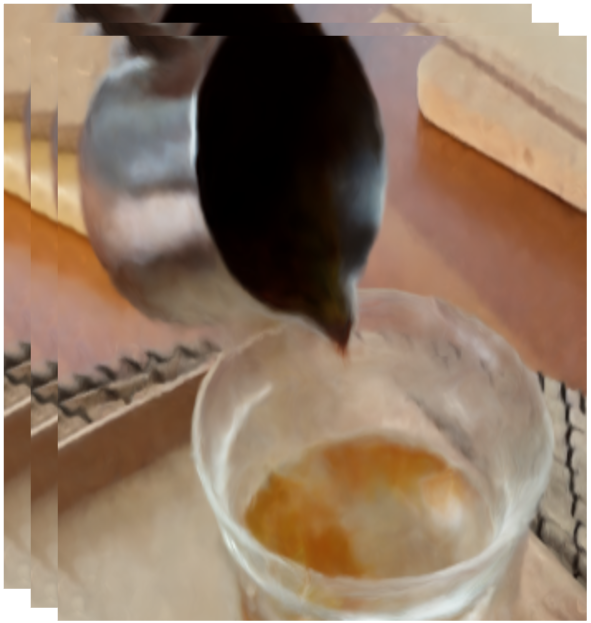} \\
    TiNeuVox \cite{tineuvox} \\
    Render time: $\SI{7}{\second}$
    \end{minipage}
    \begin{minipage}{0.15\textwidth}
    \centering
    \includegraphics[width=0.99\textwidth]{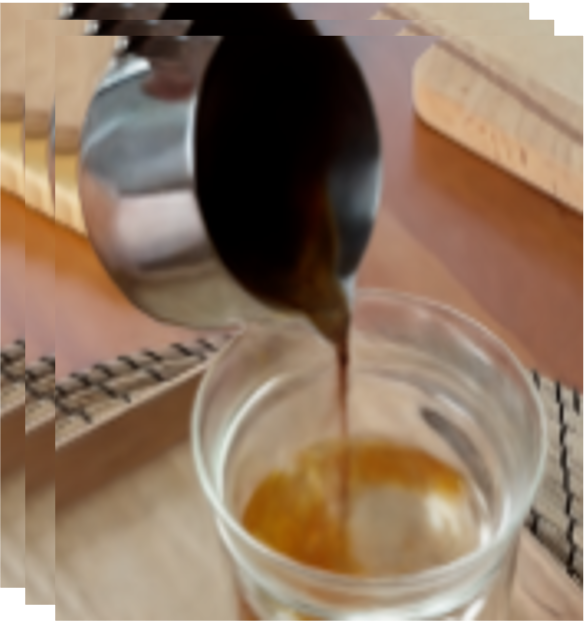} \\
    Ours \\
    Render time: $\SI{0.1}{\second}$
    \end{minipage}
    \begin{minipage}{0.28\textwidth}
    \centering
    \includegraphics[width=0.99\textwidth]{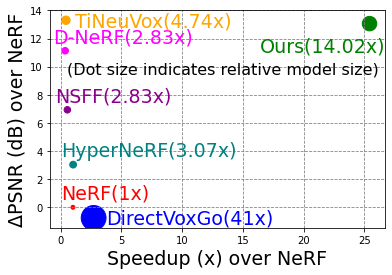} \\
    \emph{}
    \end{minipage}
    \captionof{figure}{Our proposed DyLiN for dynamic 3D scene rendering achieves higher quality than its HyperNeRF teacher model and the state-of-the-art TiNeuVox model, while being an order of magnitude faster.
    Right: DyLiN is of moderate storage size (shown by dot radii).
    For each method, the relative improvement in Peak Signal-to-Noise Ratio over NeRF ($\Delta\text{PSNR}$) is measured for the best-performing scene.
    }
    \label{fig:fancy}
\end{center}%
}]

\begin{abstract}
Light Field Networks, the re-formulations of radiance fields to oriented rays, are magnitudes faster than their coordinate network counterparts, and provide higher fidelity with respect to representing 3D structures from 2D observations. They would be well suited for generic scene representation and manipulation, but suffer from one problem: they are limited to holistic and static scenes.
In this paper, we propose the Dynamic Light Field Network (DyLiN) method that can handle non-rigid deformations, including topological changes. We learn a deformation field from input rays to canonical rays, and  lift them into a higher dimensional space to handle discontinuities.
We further introduce CoDyLiN, which augments DyLiN with controllable attribute inputs.
We train both models via knowledge distillation from pretrained dynamic radiance fields.
We evaluated DyLiN using both synthetic and real world datasets that include various non-rigid deformations.
DyLiN qualitatively outperformed and quantitatively matched state-of-the-art methods in terms of visual fidelity, while being $25 - 71\times$ computationally faster.
We also tested CoDyLiN on attribute annotated data and it surpassed its teacher model.
Project page: \url{https://dylin2023.github.io}.

\end{abstract}


\section{Introduction}
\label{sec:intro}
Machine vision has made tremendous progress with respect to reasoning about 3D structure using 2D observations. Much of this progress can be attributed to the emergence of coordinate networks \cite{chen2019net,mescheder2019occupancy,park2019deepsdf}, such as Neural Radiance Fields (NeRF) \cite{mildenhall2021nerf} and its variants \cite{barron2021mip,mildenhall2022nerf,wang2021neus,martin2021nerf}. They provide an object agnostic representation for 3D scenes and can be used for high-fidelity synthesis for unseen views.
While NeRFs mainly focus on static scenes, a series of works\cite{pumarola2021d,tretschk2021non,gafni2021dynamic,park2021nerfies} extend the idea to dynamic cases via additional components that map the observed deformations to a canonical space, supporting moving and shape-evolving objects. It was further shown that by lifting this canonical space to higher dimensions the method can handle changes in scene topology as well \cite{park2021hypernerf}.

However, the applicability of NeRF models is considerably limited by their computational complexities.
From each pixel, one typically casts a ray from that pixel, and numerically integrates the radiance and color densities computed by a Multi-Layer Perceptron (MLP) across the ray, approximating the pixel color. Specifically, the numerical integration involves sampling hundreds of points across the ray, and evaluating the MLP at all of those locations.
Several works have been proposed for speeding up static NeRFs.
These include employing a compact 3D representation structure \cite{liu2020neural,yu2021plenoctrees,fridovich2022plenoxels}, breaking up the MLP into multiple smaller networks \cite{rebain2021derf,reiser2021kilonerf}, leveraging depth information \cite{neff2021donerf,deng2022depth}, and using fewer sampling points \cite{lindell2021autoint,neff2021donerf,xu2022point}.
Yet, these methods still rely on integration and suffer from sampling many points, making them prohibitively slow for real-time applications.
Recently, Light Field Networks (LFNs) \cite{sitzmann2021lfns} proposed replacing integration with a direct ray-to-color regressor, trained using the same sparse set of images, requiring only a single forward pass.
R2L \cite{wang2022r2l} extended LFNs to use a very deep residual architecture, trained by distillation from a NeRF teacher model to avoid overfitting.
In contrast to static NeRF acceleration, speeding up dynamic NeRFs is a much less discussed problem in the literature.
This is potentially due to the much increased difficulty of the task, as one also has to deal with the high variability of motion.
In this direction, \cite{tineuvox,wang2022fourier} greatly reduce the training time by using well-designed data structures, but their solutions still rely on integration.
LFNs are clearly better suited for acceleration, yet, to the best of our knowledge, no works have attempted extending LFNs to the dynamic scenario.

In this paper, we propose 2 schemes extending LFNs to dynamic scene deformations, topological changes and controllability.
First, we introduce DyLiN, by incorporating a deformation field and a hyperspace representation to deal with non-rigid transformations, while distilling knowledge from a pretrained dynamic NeRF.
Afterwards, we also propose CoDyLiN, via adding controllable input attributes, trained with synthetic training data generated by a pretrained Controllable NeRF (CoNeRF) \cite{kania2022conerf} teacher model.
To test the efficiencies of our proposed schemes, we perform empirical experiments on both synthetic and real datasets.
We show that our DyLiN achieves better image quality and an order of magnitude faster rendering speed than its original dynamic NeRF teacher model and the state-of-the-art TiNeuVox \cite{tineuvox} method.
Similarly, we also show that CoDyLiN outperforms its CoNeRF teacher.
We further execute ablation studies to verify the individual effectiveness of different components of our model.
Our methods can be also understood as accelerated versions of their respective teacher models, and we are
not aware of any prior works that attempt speeding up CoNeRF.
\\Our contributions can be summarized as follows:
\begin{itemize}
\item We propose DyLiN, an extension of LFNs that can handle dynamic scenes with topological changes. DyLiN achieves this through non-bending ray deformations, hyperspace lifting for whole rays, and knowledge distillation from dynamic NeRFs.
\item 
We show that DyLiN achieves state-of-the-art results on both synthetic and real-world scenes, while being an order of magnitude faster than the competition.
We also include an ablation study to analyze the contributions of our model components.
\item We introduce CoDyLiN, further extending our DyLiN to handle controllable input attributes.
\end{itemize}

\section{Related Works}
\label{sec:formatting}
\paragraph{Dynamic NeRFs.}
NeRFs have demonstrated impressive performances in novel view synthesis for static scenes.
Extending these results to dynamic (deformable) domains has sparked considerable research interest \cite{pumarola2021d,tretschk2021non,gafni2021dynamic,park2021nerfies,park2021hypernerf}.
Among these works, the ones that most closely resemble ours are D-NeRF \cite{pumarola2021d} and HyperNeRF \cite{park2021hypernerf}.
D-NeRF uses a translational deformation field with temporal positional encoding.
HyperNeRF introduces a hyperspace representation, allowing topological variations to be effectively captured.
Our work expands upon these works, as we propose DyLiN, a similar method for LFNs. We use the above dynamic NeRFs as pretrained teacher models for DyLiN, achieving better fidelity with orders of magnitude shorter rendering times.

\paragraph{Accelerated NeRFs.}
The high computational complexity of NeRFs has motivated several follow-up works on speeding up the numerical integration process.
The following first set of works are restricted to static scenarios.
NSVF \cite{liu2020neural} represents the scene with a set of voxel-bounded MLPs organized in a sparse voxel octree, allowing voxels without relevant content to be skipped.
KiloNeRF \cite{reiser2021kilonerf} divides the scene into a grid and trains a tiny MLP network for each cell within the grid, saving on pointwise evaluations.
AutoInt \cite{lindell2021autoint} reduces the number of point samples for each ray using learned partial integrals.
In contrast to the above procedures, speeding up dynamic NeRFs is much less discussed in the literature, as there are only 2 papers published on this subject.
Wang \textit{et al.} \cite{wang2022fourier} proposed a method based on Fourier plenoctrees for real-time dynamic rendering, however, the technique requires an expensive rigid scene capturing setup.
TiNeuVox \cite{tineuvox} reduces training time by augmenting the MLP with time-aware voxel features and a tiny deformation network, while using a multi-distance interpolation method to model temporal variations.
Interestingly, all of the aforementioned methods suffer from sampling hundreds of points during numerical integration, and none of them support changes in topology, whereas our proposed DyLiN excels from both perspectives.

\paragraph{Light Field Networks (LFNs).}
As opposed to the aforementioned techniques that accelerate numerical integration within NeRFs, some works have attempted completely replacing numerical integration with direct per-ray color MLP regressors called Light Field Networks (LFNs).
Since these approaches accept rays as inputs, they rely heavily on the ray representation.
Several such representations exist in the literature.
Plenoptic functions \cite{bergen1991plenoptic,adelson1992plenoptic} encode 3D rays with 5D representations, i.e., a 3D point on a ray and 2 axis-angle ray directions.
Light fields \cite{gortler1996lumigraph,levoy1996light} use 4D ray codes most commonly through two-plane parameterization: given 2 parallel planes, rays are encoded by the 2D coordinates of the 2 ray-plane intersection points.
Sadly, these representations are either discontinuous or cannot represent the full set of rays.
Recently, Sitzmann \textit{et al.} \cite{sitzmann2021lfns} advocate for the usage of the 6D Pl\"{u}cker coordinate representation, i.e., a 3D point on a ray coupled with its cross product with a 3D direction.
They argue that this representation covers the whole set of rays and is continuous.
Consequently, they feed it as input to an LFN, and additionally apply Meta-Learning across scenes to learn a multi-view consistency prior.
However, they have not considered alternative ray representations, MLP architectures or training procedures, and only tested their method on toy datasets.
R2L \cite{wang2022r2l} employs an even more effective ray encoding by concatenating few points sampled from it, and proposes a very deep (88 layers) residual MLP network for LFNs.
They resolve the proneness to overfitting by training the MLP with an abundance of synthetic images generated by a pretrained NeRF having a shallow MLP.
Interestingly, they find that the student LFN model produces significantly better rendering quality than its teacher NeRF model, while being about 30 times faster.
Our work extends LFNs to dynamic deformations, topological changes and controllability, achieving similar gains over the pretrained dynamic NeRF teacher models.

\paragraph{Knowledge Distillation.}
The process of training a student model with synthetic data generated by a teacher model is called Knowledge Distillation (KD) \cite{bucilua2006model}, and it has been widely used in the vision and language domains \cite{chen2017learning, levoy1996light,wang2020collaborative,wang2021knowledge} as a form of data augmentation.
Like R2L \cite{wang2022r2l}, we also use KD for training, however, our teacher and student models are both dynamic and more complex than their R2L counterparts.

\section{Methods}
In this section, we present our two solutions for extending LFNs.
First, in \cref{s:hypernelf}, we propose DyLiN, supporting dynamic deformations and hyperspace representations via two respective MLPs.
We use KD to train DyLiN with synthetic data generated by a pretrained dynamic NeRF teacher model.
Second, in \cref{s:conelf}, we introduce CoDyLiN, which further augments DyLiN with controllability, via lifting attribute inputs to hyperspace with MLPs, and masking their hyperspace codes for disentanglement.
In this case, we also train via KD, but the teacher model is a pretrained controllable NeRF.

\subsection{DyLiN}
\label{s:hypernelf}

\subsubsection{Network Architecture}
\label{ss:hypernelf_arch}
\begin{figure}[ht]
  \centering
   \includegraphics[width=0.99\linewidth]{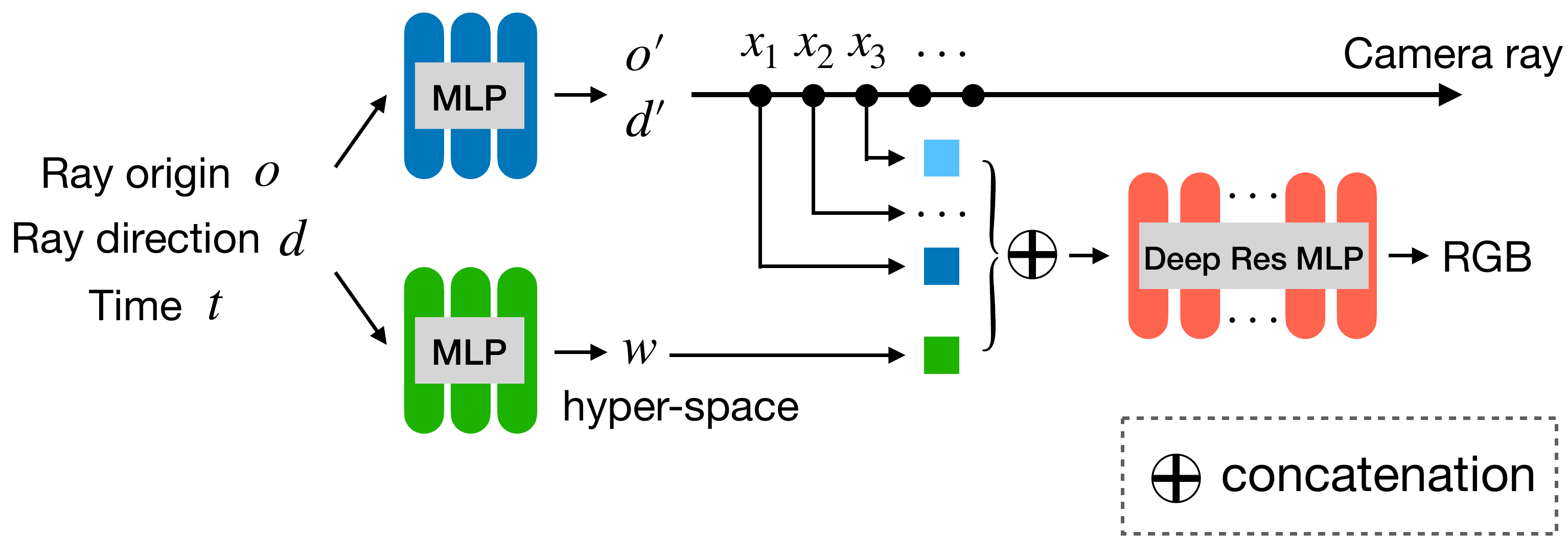}

   \caption{Schematic diagram of our proposed DyLiN architecture. We take a ray $r=(o,d)$ and time $t$ as input. We deform $r$ into $r'=(o',d')$, and sample few points $x_k$, $k=1,\dots,K$ along $r'$ to encode it (blue). In parallel, we also lift $r$ and $t$ to the hyperspace code $w$ (green), and concatenate it with each $x_k$. We use the concatenation to regress the RGB color of $r$ at $t$ directly (red).}
   \label{fig:hyperr2l}
\end{figure}

Our overall DyLiN architecture $G_\phi$ is summarized in \cref{fig:hyperr2l}.
It processes rays instead of the widely adopted 3D point inputs as follows.

Specifically, our deformation MLP $T_\omega$ maps an input ray $r=(o,d)$ to canonical space ray $r'=(o',d')$:
\begin{equation}
  (o', d') = T_\omega(o, d, t).
  \label{eq:T}
\end{equation}
Unlike the pointwise deformation MLP proposed in Nerfies \cite{park2021nerfies}, which bends rays by offsetting their points independently, our MLP outputs rays explicitly, hence no ray bending occurs.
Furthermore, after obtaining $r'$, we encode it by sampling and concatenating $K$ points along it.

Our hyperspace MLP $H_\psi$ is similar to $T_\omega$, except it outputs a hyperspace representation $w$:
\begin{equation}
  w = H_\psi(o, d, t).
  \label{eq:H}
\end{equation}
In contrast to HyperNeRF \cite{park2021hypernerf}, which predicts a hyperspace code $w$ for each 3D point, we use rays and compute a single $w$ for each ray.

Both MLPs further take the index $t$ as input to encode temporal deformations.

Once the $K$ points and $w$ are obtained, we concatenate them and feed the result into our LFN $R_\pi$, which is a deep residual color MLP regressor.
Overall, we can collect the model parameters as $\phi=[\omega,\psi,\pi]$.

Note that without our two MLPs $T_\omega$ and $H_\psi$, our DyLiN falls back to the vanilla LFN.

\subsubsection{Training Procedure}
\label{ss:hypernelf_train}
Our training procedure is composed of 3 phases.

First, we pretrain a dynamic NeRF model $F_{\theta}$ (e.g., D-NeRF \cite{pumarola2021d} or HyperNeRF \cite{park2021hypernerf}) by randomly sampling time $t$ and input ray $r$, and minimizing the Mean Squared Error (MSE) against the corresponding RGB color of monocular target video $I$:
\begin{equation}
  \min_\theta \,\mathbb{E}_{t,r=(o,d)}\left[\|F_{\theta}(o,d,t)-I(o,d,t)\|_2^2\right].
  \label{eq:L1}
\end{equation}
Recall, that $F_\theta$ is slow, as it performs numerical integration across the ray $r=(o,d)$.

Second, we employ the newly obtained $F_{\theta^*}$ as the teacher model for our DyLiN student model $G_\phi$ via KD.
Specifically, we minimize the MSE loss against the respective pseudo ground truth ray color generated by $F_{\theta^*}$ across $S$ ray samples:
\begin{equation}
  \min_\phi \,\mathbb{E}_{t,r=(o,d)}\left[\|G_\phi(o,d,t)-F_{\theta^*}(o,d,t)\|_2^2\right],
  \label{eq:L2}
\end{equation}
yielding $G_{\tilde{\phi}}$.
Note how this is considerably different from R2L \cite{wang2022r2l}, which uses a static LFN that is distilled from a static NeRF.

Finally, we initialize our student model $G_\phi$ with parameters $\tilde{\phi}$ and fine-tune it using the original real video data:
\begin{equation}
  \min_{\phi,\,\phi_0=\tilde{\phi}} \,\mathbb{E}_{t,r=(o,d)}\left[\|G_{\phi}(o,d,t)-I(o,d,t)\|_2^2\right],
  \label{eq:L3}
\end{equation}
obtaining $\phi^*$.


\subsection{CoDyLiN}
\label{s:conelf}
\subsubsection{Network Architecture}
\label{ss:conelf_arch}
\begin{figure}[ht]
  \centering
   \includegraphics[width=0.99\linewidth]{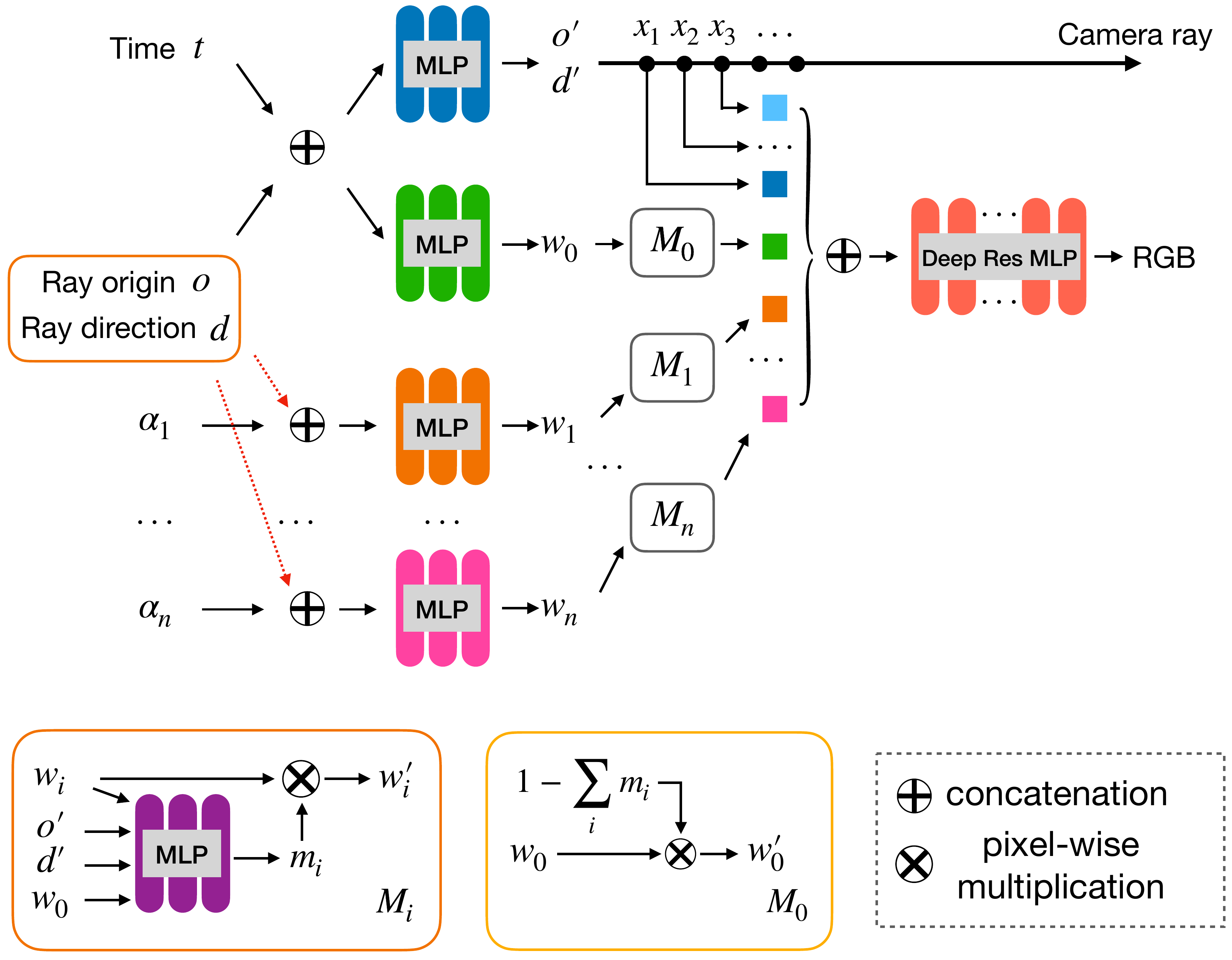}

   \caption{Schematic diagram of our proposed CoDyLiN architecture. We augment our DyLiN (blue, green, red) by introducing scalar attribute inputs $\alpha_i\in[-1,1]$, $i=1,\dots,n$ and lifting them to their respective hyperspace codes $w_i$ (orange, \dots, pink MLPs). Next, $M_i$ disentangles $w_i$ from $w_j$, $j\neq i$ by masking it into $w_i'$ (orange, \dots, pink boxes and bottom insets). We concatenate the sampled points $x_k$, $k=1,\dots,K$ with the $w_i'$, $i=1,\dots,n$ and predict the RGB color corresponding to the inputs (red).
   Arrows from $(o', d')$ and $w_0$ to $M_i$ are omitted from the top figure for simplicity.
   Compare this with \cref{fig:hyperr2l}.}
   \label{fig:cor2l}
\end{figure}

We further demonstrate that our DyLiN architecture from \cref{ss:hypernelf_arch} can be extended to the controllable scenario using attribute inputs with hyperspace MLPs and attention masks.
Our proposed CoDyLiN network $Q_\tau$ is depicted in \cref{fig:cor2l}.

Specifically, we start from DyLiN $G_\phi$ and add scalar inputs $\alpha_i\in[-1,1]$, $i=1,\dots,n$ next to $o,d,t$.
Intuitively, these are given strength values for specific local attributes, which can be interpolated continuously.
$n$ is the total number of attributes.

Each $\alpha_i$ is then processed independently with its own hyperspace MLP $H_{i,\psi_i}$ to yield the hyperspace code $w_i$:
\begin{equation}
    w_i = H_{i,\psi_i}(o,d,t).
\end{equation}

Next, we include mask MLP regressors $M_{i,\rho_i}$ to generate scalar attention masks $\hat{m}_i\in[0,1]$ for each $w_i$ (including $w_0=w$):
\begin{equation}
\begin{alignedat}{1}
    \hat{m}_i &= M_{i,\rho_i}(w_i,w,o,d),\\
    \hat{m}_0 &= 1-\sum_{i=1}^{n}{\hat{m}_i},\\
    w_i' &= \hat{m}_i \cdot w_i, \quad i=0,\dots,n,
\end{alignedat}
\end{equation}
This helps the architecture to spatially disentangle (i.e., localize) the effects of attributes $\alpha_i$, while $\hat{m}_0$ can be understood as the space not affected by any attributes.

Finally, we sample $K$ points on the ray similarly to \cref{ss:hypernelf_arch}, concatenate those with the $w_i'$ vectors, and process the result further with LFN $R_\pi$.
Again, we can use a shorthand for the parameters: $\tau=[\omega,\psi,\psi_1,\dots,\psi_n,\rho_1,\dots,\rho_n,\pi]$.

Observe that without our MLPs $H_{i,\psi_i}$, $M_{i,\rho_i}$, $i=1,\dots,n$, our CoDyLiN reverts to our simpler DyLiN.
Different from CoNeRF \cite{kania2022conerf}, we process rays instead of points, and use the $\alpha_i$ as inputs instead of targets.

\subsubsection{Training Procedure}
\label{ss:conelf_train}
Akin to \cref{ss:hypernelf_train}, we split training into pretraining and distillation steps, but omit fine-tuning.

First, we pretrain a CoNeRF model $E_\nu$ \cite{kania2022conerf} by randomly sampling $(t,r,i)$, against 3 ground truths: ray color, attribute values $\alpha_i$ and 2D per-attribute masks $m_{2D,i}$.
This yields us $E_{\nu^*}$.
For brevity, we omit the details of this step, and kindly forward the reader to Section 3 in \cite{kania2022conerf}.

Second, we distill from our teacher CoNeRF model $E_{\nu^*}$ into our student CoDyLiN $Q_\tau$ by randomly sampling $t,r,\alpha_1,\dots,\alpha_n$, and minimizing the MSE against 2 pseudo ground truths, i.e., ray colors and 2D masks $\bar{m}_{2D,i}$:
\begin{multline}
  \min_\tau \,\mathbb{E}_{t,r=(o,d)}\biggl[\|Q_\tau(o,d,t,\alpha_{1:n})-\bar{E}_{\nu^*}(o,d,t,\alpha_{1:n})\|_2^2 \\ 
  +\lambda_m \cdot \sum_{i=0}^{n}{\|\hat{m}_i(o,d,t,\alpha_{i})-\bar{m}_{2D}(o,d,t,\alpha_{1:n})_{i}\|_2^2}\biggr],
  \label{eq:L5}
\end{multline}
where $\bar{E}_\nu$ is identical to $E_\nu$ except for taking $\alpha_{1:n}=[\alpha_1,\dots,\alpha_n]$ as input and outputting the masks $\bar{m}_{2D,i}$, $i=0,\dots,n$.
We denote the result of the optimization as $Q_{\tau^*}$.


We highlight that our teacher and student models are both controllable in this setup.

\section{Experimental Setup}




\subsection{Datasets}
To test our hypotheses, we performed experiments on three types of dynamic scenes: synthetic, real and real controllable. \\
\textbf{Synthetic Scenes.}
We utilized the synthetic $360^{\circ}$ dynamic dataset introduced by \cite{pumarola2021d}, which contains 8 animated objects with complicated geometry and realistic non-Lambertian materials.
Each dynamic scene consists of $50$ to $200$ training images and $20$ testing images.
We used $400 \times 400$ image resolution.
We applied D-NeRF \cite{pumarola2021d} as our teacher model with the publicly available pretrained weights.\\
\textbf{Real Scenes.}
We collected real dynamic data from $2$ sources.
First, we  utilized 5 topologically varying scenes provided by \cite{park2021hypernerf} (Broom, 3D Printer, Chicken, Americano and Banana), which were captured by a rig encompassing a pole with two Google Pixel 3 phones rigidly attached roughly $\SI{16}{\centi\metre}$ apart.
Second, we collected human facial videos using an iPhone 13 Pro camera.
We rendered both sets at $960 \times 540$ image resolution.
We pretrained a HyperNeRF \cite{park2021hypernerf} teacher model from scratch for each scene.\\
\textbf{Real Controllable Scenes.}
We borrowed 2 real controllable scenes from \cite{kania2022conerf} (closing/opening eyes/mouth, and transformer), which are captured either with a Google Pixel 3a or an Apple iPhone 13 Pro, and contain annotations over various attributes.
We applied image resolution of $480 \times 270$ pixels.
We pretrained a CoNeRF \cite{kania2022conerf} teacher model from scratch per scene.

\subsection{Settings}
Throughout our experiments, we use the settings listed below, many of which follow \cite{wang2022r2l}.

In order to retain efficiency, we define $T_\omega$ and $H_\psi$ to be small MLPs, with $T_\omega$ consisting of $7$ layers of $128$ units with $r'\in\mathbb{R}^6$, and $H_\psi$ having $6$ layers of $64$ units with $w\in\mathbb{R}^8$.
Then, we use $K=16$ sampled points to represent rays, where sampling is done randomly during training and evenly spaced during inference.

Contrary to $T_\omega$ and $H_\psi$, our LFN $R_\pi$ is a very deep residual color MLP regressor, containing $88$ layers with $256$ units per layer, in order to have enough capacity to learn the video generation process.

We generate rays within \cref{eq:L1,eq:L2,eq:L3,eq:L5} by sampling ray origins $o=(x_o, y_o, z_o)$ and normalized directions $d=(x_d,y_d,z_d)$ randomly from the uniform distribution $U$ as follows:
\begin{alignat}{2}
  x_o &\sim U(x_o^{min}, x_o^{max}), \quad & x_d &\sim U(x_d^{min}, x_d^{max}),\\
  y_o &\sim U(y_o^{min}, y_o^{max}), \quad & y_d &\sim U(y_d^{min}, y_d^{max}),\\
  z_o &\sim U(z_o^{min}, z_o^{max}), \quad & z_d &\sim U(z_d^{min}, z_d^{max}),
  \label{eq:U}
\end{alignat}
where the $min,max$ bounds of the 6 intervals are inferred from the original training video.
In addition to uniform sampling, we also apply the hard example mining strategy suggested in \cite{wang2022r2l} to focus on fine-grained details.
We used $S=\SI{10000}{}$ training samples during KD in \eqref{eq:L2}.

Subsequently, we also randomly sample time step $t$ uniformly from the unit interval: $t \sim U(0,1)$.

Optionally, for our CoDyLiN experiments, we define each $H_{i,\psi_i}$ to be a small MLP having $5$ layers of $128$ units with $w_i\in\mathbb{R}^8$.
During training, we uniformly sample attributes within $[-1,1]$: $\alpha_i \sim U(-1, 1)$, and let $\lambda_m=0.1$.

During training, we used Adam \cite{kingma2014adam} with learning rate $\SI{5e-4}{}$ and batch size $\SI{4096}{}$.

We performed all experiments on single NVIDIA A100 GPUs.


\subsection{Baseline Models}
For testing our methods, we compared quality and speed against several baseline models, including NeRF \cite{mildenhall2021nerf}, NV \cite{Lombardi:2019}, NSFF \cite{li2020neural}, Nerfies \cite{park2021nerfies}, HyperNeRF \cite{park2021hypernerf}, two variants of TiNeuVox \cite{tineuvox}, DirectVoxGo \cite{SunSC22}, Plenoxels \cite{fridovich2022plenoxels}, T-NeRF and D-NeRF \cite{pumarola2021d}, as well as CoNeRF \cite{kania2022conerf}.

\begin{figure}[htb]
     \centering
     \begin{subfigure}[b]{0.45\textwidth}
         \centering
         \includegraphics[width=0.99\textwidth]{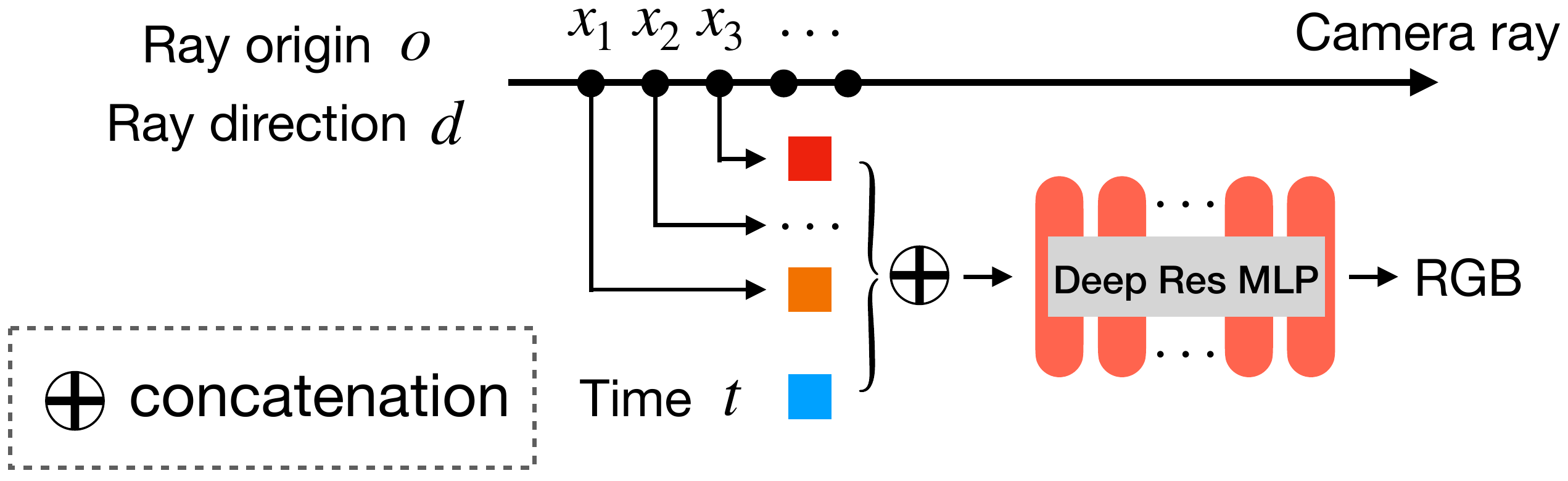}
         \caption{}
         \label{fig:T-R2L}
     \end{subfigure}
     \hfill
     \begin{subfigure}[b]{0.45\textwidth}
         \centering
         \includegraphics[width=0.99\textwidth]{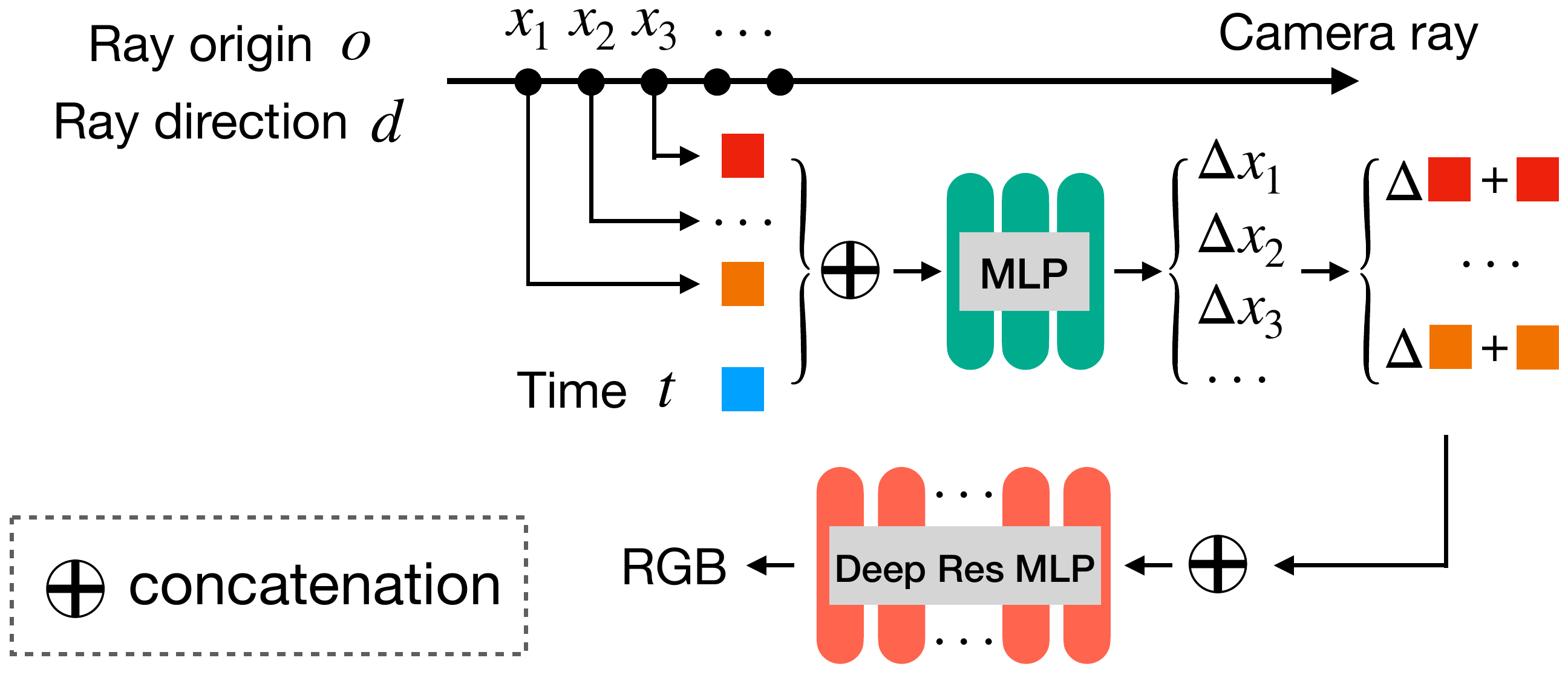}
         \caption{}
         \label{fig:D-R2L}
     \end{subfigure}
        \caption{Our two ablated baseline models, omiting components of our DyLiN. (a)~Without our two proposed MLPs. (b)~Pointwise deformation MLP only, predicting offsets jointly.}
        \label{fig:TD-R2L}
\end{figure}

In addition, we performed an ablation study by comparing against 2 simplified versions of our DyLiN architecture.
First, we omitted both of our deformation and hyperspace MLPs and simply concatenated the time step $t$ to the sampled ray points (essentially resulting in a dynamic R2L).
This method is illustrated in \cref{fig:T-R2L}.
Second, we employed a pointwise deformation MLP ($5$ layers of $256$ units) inspired by \cite{pumarola2021d}, which deforms points along a ray by predicting their offsets jointly, i.e., it can bend rays.
This is contrast to our DyLiN, which deforms rays explicitly without bending and also applies a hyperspace MLP.
This scheme is depicted in \cref{fig:D-R2L}.
In both baselines, the deep residual color MLP regressors were kept intact.
Next, we also tested the effects of our fine-tuning procedure from \eqref{eq:L3} by training all of our models both with and without it.
Lastly, we assessed the dependences on the number of sampled points along rays $K$ and on the number of training samples $S$ during KD in \eqref{eq:L2}.


\subsection{Evaluation Metrics}
For quantitatively evaluating the quality of generated images, we calculated the Peak Signal-to-Noise Ratio (PSNR) \cite{hore2010image} in decibels ($\si{\deci\bel}$), the Structural Similarity Index (SSIM) \cite{wang2004image, odena2017conditional}, the Multi-Scale SSIM (MS-SSIM) \cite{wang2003multiscale} and the Learned Perceptual Image Patch Similarity (LPIPS) \cite{zhang2018unreasonable} metrics.
Intuitively, PSNR is a pixelwise score, while SSIM and MS-SSIM also take pixel correlations and multiple scales into account, respectively, yet all of these tend to favor blurred images. 
LPIPS compares deep neural representations of images and is much closer to human perception, promoting semantically better and sharper images.

Furthermore, for testing space and time complexity, we computed the storage size of parameters in megabytes (MB) and measured the wall-clock time in milliseconds (ms) while rendering the synthetic Lego scene with each model.


\section{Results}
\subsection{Quantitative Results}
\cref{tab:synth} and \cref{tab:real} contain our quantitative results for reconstruction quality on synthetic and real dynamic scenes, accordingly.
We found that among prior works, TiNeuVox-B performed the best on synthetic scenes with respect to each metric. On real scenes, however, NSFF took the lead. Despite having strong metrics, NSFF is qualitatively poor and slow.
Surprisingly, during ablation, even our most basic model (DyLiN without the two MLPs from \cref{fig:T-R2L}) could generate perceptually better looking images than TiNeuVox-B, thanks to the increased training dataset size via KD.
Incorporating the MLPs $T_\omega$ and $H_\psi$ into the model each improved results slightly.
Interestingly, fine-tuning on real data as in \eqref{eq:L3} gave a substantial boost.
In addition, our relative PSNR improvement over the teacher model (\cref{tab:synth}=$+\SI{1.93}{\deci\bel}$, up to $+\SI{3.16}{\deci\bel}$ per scene; \cref{tab:real}=$+\SI{2.7}{\deci\bel}$, up to $+\SI{13.14}{\deci\bel}$) is better than that of R2L \cite{wang2022r2l} ($+\SI{1.4}{\deci\bel}$, up to $+\SI{2.8}{\deci\bel}$).

\begin{table}[htb]
\centering
\caption{Quantitative results on synthetic dynamic scenes. Notations: Multi-Layer Perceptron (MLP), PD (pointwise deformation), FT (fine-tuning). We utilized D-NeRF as the teacher model for our DyLiNs. The winning numbers are highlighted in bold.}
\resizebox{\columnwidth}{!}{%

\begin{tabular}{lccc}
\toprule
Method                                 & PSNR$\uparrow$ & SSIM$\uparrow$ & LPIPS$\downarrow$ \\ \midrule
NeRF\cite{mildenhall2021nerf}          & 19.00          & 0.8700           & 0.1825              \\
DirectVoxGo\cite{SunSC22}              & 18.61          & 0.8538           & 0.1688              \\
Plenoxels\cite{fridovich2022plenoxels} & 20.24          & 0.8688           & 0.1600              \\
T-NeRF\cite{pumarola2021d}             & 29.51          & 0.9513           & 0.0788              \\
D-NeRF\cite{pumarola2021d}             & 30.50          & 0.9525           & 0.0663              \\
TiNeuVox-S\cite{tineuvox}              & 30.75          & 0.9550           & 0.0663              \\
TiNeuVox-B\cite{tineuvox}              & \textbf{32.67}          & 0.9725           & 0.0425              \\ 
\midrule
DyLiN, w/o two MLPs, w/o FT (ours)                & 31.16          & 0.9931           & 0.0281                \\
DyLiN, w/o two MLPs (ours)           & 32.07          & 0.9937           & 0.0196                  \\
DyLiN, PD MLP only, w/o FT (ours)                & 31.26          & 0.9932           & 0.0279                  \\
DyLiN, PD MLP only (ours)           & 31.24          & 0.9940           & 0.0189                  \\
DyLiN, w/o FT (ours)            & 31.37          & 0.9933           & 0.0275                  \\
DyLiN (ours)       & 32.43          & \textbf{0.9943}           & \textbf{0.0184} \\                 
\bottomrule
\end{tabular}%
}
\label{tab:synth}
\end{table}

\begin{table}[ht]
\centering
\caption{Quantitative results on real dynamic scenes. Notations: Multi-Layer Perceptron (MLP), PD (pointwise deformation), FT (fine-tuning). We utilized HyperNeRF as the teacher model for our DyLiNs. The winning numbers are highlighted in bold.}
\resizebox{\columnwidth}{!}{%
\begin{tabular}{lcc}
\toprule
Method                                                  & PSNR$\uparrow$ & MS-SSIM$\uparrow$\\ \midrule
NeRF\cite{mildenhall2021nerf}          &  20.1         &       0.745                    \\
NV\cite{Lombardi:2019}                &   16.9        &      0.571                     \\
NSFF\cite{li2020neural}              &   \textbf{26.3}        &    \textbf{0.916}                       \\
Nerfies\cite{park2021nerfies}       &  22.2         &     0.803                      \\
HyperNeRF\cite{park2021hypernerf}   & 22.4          &       0.814                    \\
TiNeuVox-S\cite{tineuvox}              &  23.4         &       0.813                    \\
TiNeuVox-B\cite{tineuvox}              &    24.3       &   0.837                       \\
\midrule
DyLiN, w/o two MLPs, w/o FT (ours)                                 &   23.8             &     0.882                              \\
DyLiN, w/o two MLPs (ours)                            &   24.2             &     0.894                              \\
DyLiN, PD MLP only, w/o FT (ours)                                 &   23.9             &     0.885                              \\
DyLiN, PD MLP only (ours)                            &   24.6             &     0.903                              \\
DyLiN, w/o FT (ours)                             &   24.0             &     0.886                              \\
DyLiN (ours)                         &   25.1             &     0.910                            \\
\bottomrule
\end{tabular}%
}
\label{tab:real}
\end{table}

\begin{table}[ht]
\centering
\caption{Quantitative results for space and time complexity on the synthetic Lego scene. Notations: Multi-Layer Perceptron (MLP), PD (pointwise deformation), FT (fine-tuning).}
\resizebox{\columnwidth}{!}{%
\begin{tabular}{lcc}
\toprule
                                & Storage  & Wall-clock\\
Method & (MB) & time (ms)\\
\midrule
NeRF\cite{mildenhall2021nerf}          &  \phantom{00}5.00          &  2,950.0                     \\
DirectVoxGo\cite{SunSC22}              &  205.00         &   1,090.0                       \\
Plenoxels\cite{fridovich2022plenoxels} &  717.00         &  \phantom{0,0}50.0                       \\
NV\cite{Lombardi:2019}                 &  439.00  &         \phantom{0,0}74.9                               \\
D-NeRF\cite{pumarola2021d}             &   \phantom{00}4.00        & 8,150.0                         \\
NSFF\cite{li2020neural}                &\phantom{0}14.17        &   5,450.0                       \\
HyperNeRF\cite{park2021hypernerf}      &   \phantom{0}15.36       & 2,900.0                  \\
TiNeuVox-S\cite{tineuvox}              &   \phantom{0}23.70        & 3,280.0                          \\
TiNeuVox-B\cite{tineuvox}              &   \phantom{0}23.70        & 6,920.0                     \\ \midrule
DyLiN, w/o two MLPs, w/o FT (ours)                                 &     \phantom{0}68.04           &   \phantom{0,}115.4                         \\
DyLiN, w/o two MLPs (ours)                            &     \phantom{0}68.04           &    \phantom{0,}115.4                                \\
DyLiN, PD MLP only, w/o FT (ours)                                 &      \phantom{0}72.60          &     \phantom{0,}115.7                               \\
DyLiN, PD MLP only (ours)                            &      \phantom{0}72.60          &  \phantom{0,}115.7                              \\
DyLiN, w/o FT (ours)                             &      \phantom{0}70.11          &   \phantom{0,}116.0                                \\
DyLiN (ours)                        &      \phantom{0}70.11          &    \phantom{0,}116.0          \\
\bottomrule
\end{tabular}%
}
\label{tab:synth_spacetime}
\end{table}

\cref{tab:synth_spacetime} shows quantitative results for space and time complexity on the synthetic Lego scene.
We found that there is a trade-off between the two metrics, as prior works are typically optimized for just one of those.
In contrast, all of our proposed DyLiN variants settle at the golden mean between the two extremes.
When compared to the strongest baseline TiNeuVox-B, our method requires $3$ times as much storage but is nearly 2 orders of magnitude faster.
Plenoxels and NV, the only methods that require less computation than ours, perform much worse in quality.

\begin{figure}[ht]
\centering
\begin{subfigure}{0.45\textwidth}
  \centering
  \includegraphics[width=0.99\linewidth]{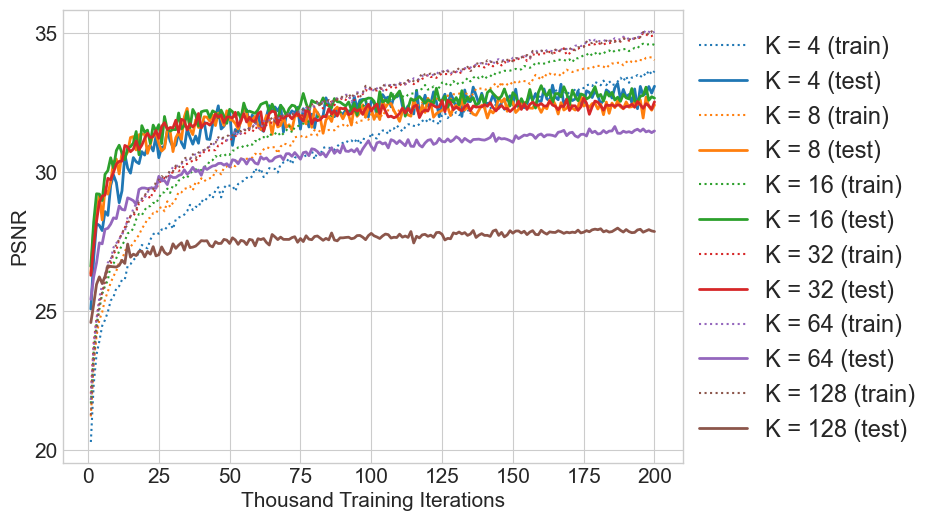}
  \caption{}
  \label{fig:ablation_sampled_point}
\end{subfigure}%
\hfill
\begin{subfigure}{0.45\textwidth}
  \centering
  \includegraphics[width=0.99\linewidth]{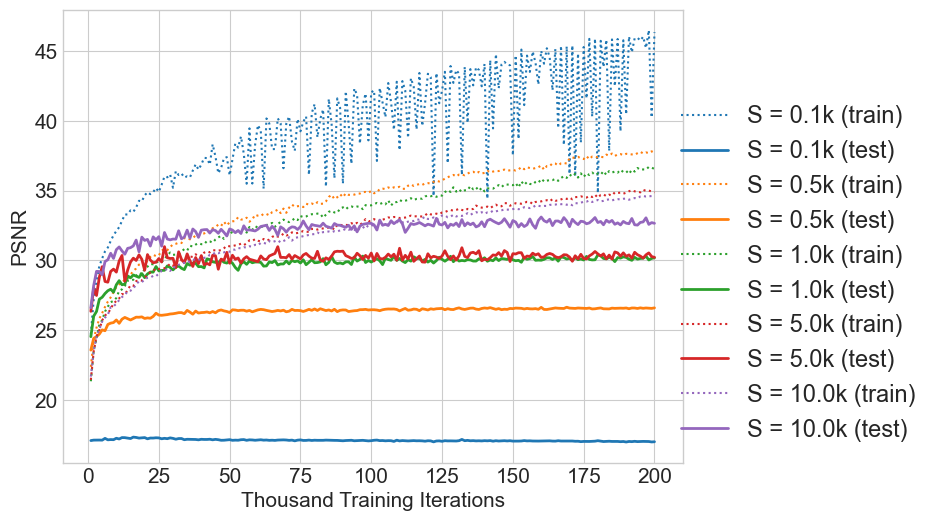}
  \caption{}
  \label{fig:ablation_pseudo_samples}
\end{subfigure}
\caption{Quantitative results for ablation on the synthetic Standup scene. (a)~Dependence on the number of sampled points $K$ across ray $r'$. (b)~Dependence on the number of training samples $S$ during Knowledge Distillation (KD).}
\label{fig:ablation}
\end{figure}

\cref{fig:ablation} reports quantitative ablation results for dependencies on the number of sampled points per ray $K$ and on the number of training samples during KD $S$, performed on the synthetic Standup scene.
For dependence on $K$ (\cref{fig:ablation_sampled_point}), we found that there were no significant differences between test set PNSR scores for $K\in\{4,8,16,32\}$, while we encountered overfitting for $K\in\{64,128\}$.
This justified our choice of $K=16$ for the rest of our experiments.
Regarding the effect of $S$ (\cref{fig:ablation_pseudo_samples}), overfitting occured for smaller sample sizes including $S\in\{100;500;\SI{1000}{};\SI{5000}\}$.
The test and training set PSNR scores were much closer for $S=\SI{10000}{}$, validating our general setting.

Our controllable numerical results are collected in \cref{tab:control}.
In short, our CoDyLiN was able to considerably outperform CoNeRF with respect to MS-SSIM and speed.

\begin{table}[ht]
\centering
\caption{Quantitative results on real controllable scenes. We utilized CoNeRF as the teacher model for our CoDyLiN. The winning numbers are highlighted in bold.}
\resizebox{\columnwidth}{!}{%
\begin{tabular}{lcccccc}
\toprule
& \multicolumn{3}{c}{Eyes/Mouth}  & \multicolumn{3}{c}{Transformer} \\
\cmidrule(lr){2-4} \cmidrule(lr){5-7}
 &  &   & Wall-clock & &  & Wall-clock \\ 
Method & PSNR$\uparrow$ & MS-SSIM$\uparrow$ & time ($\si{\milli\second}$) & PSNR$\uparrow$ & MS-SSIM$\uparrow$ & time ($\si{\milli\second}$)\\
 \midrule
CoNeRF\cite{kania2022conerf}          &      \textbf{21.4658}     &   0.7458    & 6230.0&       23.0319  & 0.8878 & 4360.0           \\
CoDyLiN (ours)                        &       21.4655    &   \textbf{0.9510}   &  \phantom{0}\textbf{116.3} &      \textbf{23.5882} & \textbf{0.9779} & \phantom{0}\textbf{116.0}                   
\\
\bottomrule
\end{tabular}%
}
\label{tab:control}
\end{table}

\subsection{Qualitative Results}
\begin{figure*}[!ht]
     \centering
     \begin{subfigure}[b]{0.15\textwidth}
         \centering
         \includegraphics[width=0.99\textwidth]{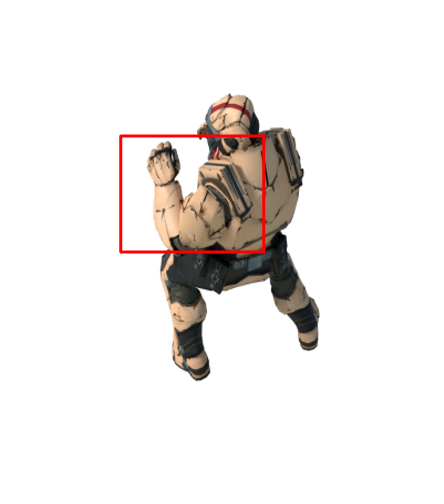}
         \caption*{\textbf{Hook}}
         \label{fig:hook_box}
     \end{subfigure}     
     \setcounter{subfigure}{0}
     \begin{subfigure}[b]{0.15\textwidth}
         \centering
         \includegraphics[width=0.99\textwidth]{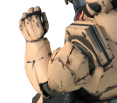}
         \caption{Ground Truth}
         \label{fig:hook_gt}
     \end{subfigure}
     \begin{subfigure}[b]{0.15\textwidth}
         \centering
         \includegraphics[width=0.99\textwidth]{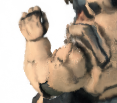}
         \caption{D-NeRF \cite{pumarola2021d}}
         \label{fig:hook_dnerf}
     \end{subfigure}
    \begin{subfigure}[b]{0.15\textwidth}
         \centering
         \includegraphics[width=0.99\textwidth]{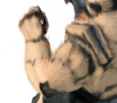}
         \caption{TiNeuVox \cite{tineuvox}}
         \label{fig:hook_tineuvox}
     \end{subfigure}
    \begin{subfigure}[b]{0.15\textwidth}
         \centering
         \includegraphics[width=0.99\textwidth]{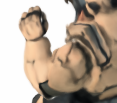}
         \caption{Ours-1}
         \label{fig:hook_ours1}
     \end{subfigure}
    \begin{subfigure}[b]{0.15\textwidth}
         \centering
         \includegraphics[width=0.99\textwidth]{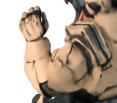}
         \caption{Ours-2}
         \label{fig:hook_ours2}
     \end{subfigure}    
     
     \begin{subfigure}[b]{0.15\textwidth}
         \centering
         \includegraphics[width=0.99\textwidth]{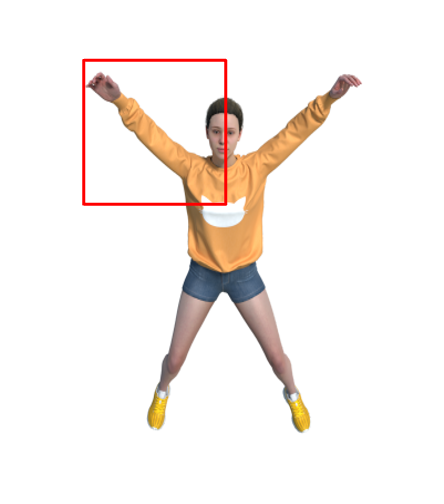}
         \caption*{\textbf{Jumping Jacks}}
         \label{fig:jumping_box}
     \end{subfigure}     
     \begin{subfigure}[b]{0.15\textwidth}
         \centering
         \includegraphics[width=0.99\textwidth]{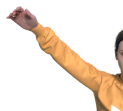}
         \caption{Ground Truth}
         \label{fig:jumping_gt}
     \end{subfigure}
     \begin{subfigure}[b]{0.15\textwidth}
         \centering
         \includegraphics[width=0.99\textwidth]{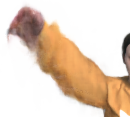}
         \caption{D-NeRF \cite{pumarola2021d}}
         \label{fig:jumping_dnerf}
     \end{subfigure}
    \begin{subfigure}[b]{0.15\textwidth}
         \centering
         \includegraphics[width=0.99\textwidth]{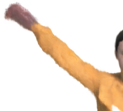}
         \caption{TiNeuVox \cite{tineuvox}}
         \label{fig:jumping_tineuvox}
     \end{subfigure}
    \begin{subfigure}[b]{0.15\textwidth}
         \centering
         \includegraphics[width=0.99\textwidth]{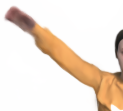}
         \caption{Ours-1}
         \label{fig:jumping_ours1}
     \end{subfigure}
    \begin{subfigure}[b]{0.15\textwidth}
         \centering
         \includegraphics[width=0.99\textwidth]{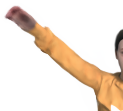}
         \caption{Ours-2}
         \label{fig:jumping_ours2}
     \end{subfigure}

        \caption{Qualitative results on synthetic dynamic scenes. We compare our DyLiN (Ours-1, Ours-2) with the ground truth, the D-NeRF teacher model and TiNeuVox.
        Ours-1 and Ours-2 were trained without and with fine-tuning on the original data, respectively.}
        \label{fig:qual_synth}
\end{figure*}

\begin{figure*}[!ht]
     \centering
     \begin{subfigure}[b]{0.15\textwidth}
         \centering
         \includegraphics[width=0.99\textwidth,height=2.03cm,keepaspectratio]{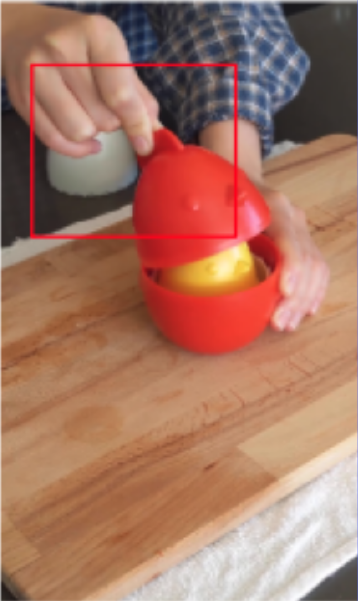}
         \caption*{\textbf{Chicken}}
         \label{fig:chicken_box}
     \end{subfigure}     
     \setcounter{subfigure}{0}
     \begin{subfigure}[b]{0.15\textwidth}
         \centering
         \includegraphics[width=0.99\textwidth]{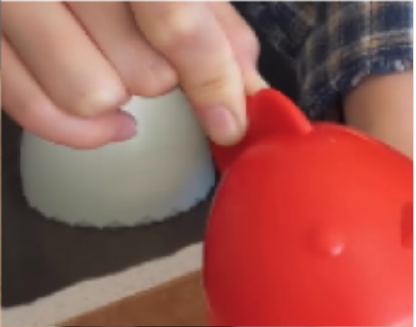}
         \caption{Ground Truth}
         \label{fig:chicken_gt}
     \end{subfigure}
     \begin{subfigure}[b]{0.15\textwidth}
         \centering
         \includegraphics[width=0.99\textwidth]{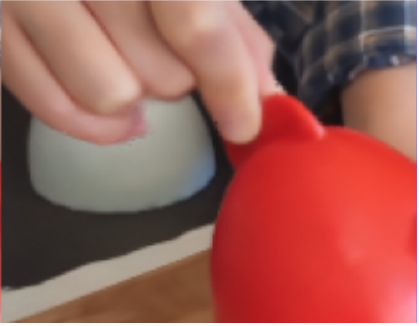}
         \caption{HyperNeRF \cite{park2021hypernerf}}
         \label{fig:chicken_hypernerf}
     \end{subfigure}
    \begin{subfigure}[b]{0.15\textwidth}
         \centering
         \includegraphics[width=0.99\textwidth]{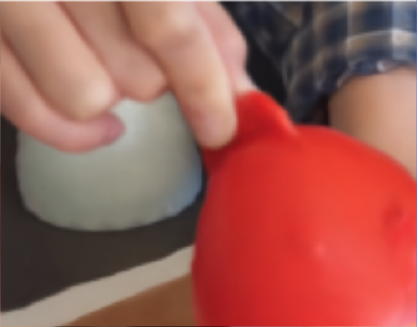}
         \caption{TiNeuVox \cite{tineuvox}}
         \label{fig:chicken_tineuvox}
     \end{subfigure}
    \begin{subfigure}[b]{0.15\textwidth}
         \centering
         \includegraphics[width=0.99\textwidth]{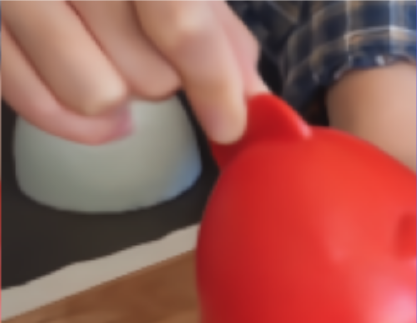}
         \caption{Ours-1}
         \label{fig:chicken_ours1}
     \end{subfigure}
    \begin{subfigure}[b]{0.15\textwidth}
         \centering
         \includegraphics[width=0.99\textwidth]{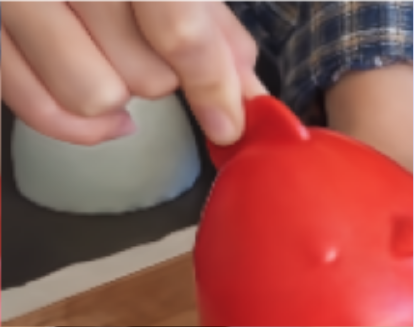}
         \caption{Ours-2}
         \label{fig:chicken_ours2}
     \end{subfigure}    

        \caption{
        Qualitative results on a real dynamic scene.
        We compare our DyLiN (Ours-1, Ours-2) with the ground truth, the HyperNeRF teacher model and TiNeuVox.
        Ours-1 and Ours-2 were trained without and with fine-tuning on the original data, respectively.}
        \label{fig:qual_real}
\end{figure*}

\cref{fig:qual_synth} and \cref{fig:qual_real} depict qualitative results for reconstruction quality on synthetic and real dynamic scenes, respectively.
Both show that our full DyLiN model generated the sharpest, most detailed images, as it was able to capture cloth wrinkles (\cref{fig:jumping_ours2}) and the eye of the chicken (\cref{fig:chicken_ours2}).
The competing methods tended to oversmooth these features.
We also ablated the effect of omitting fine-tuning (\cref{fig:jumping_ours1}, \cref{fig:chicken_ours1}), and results declined considerably.

\begin{figure*}[!htb]
     \centering
     \begin{subfigure}[b]{0.15\textwidth}
         \centering
         \includegraphics[width=0.99\textwidth,height=2.152cm,keepaspectratio]{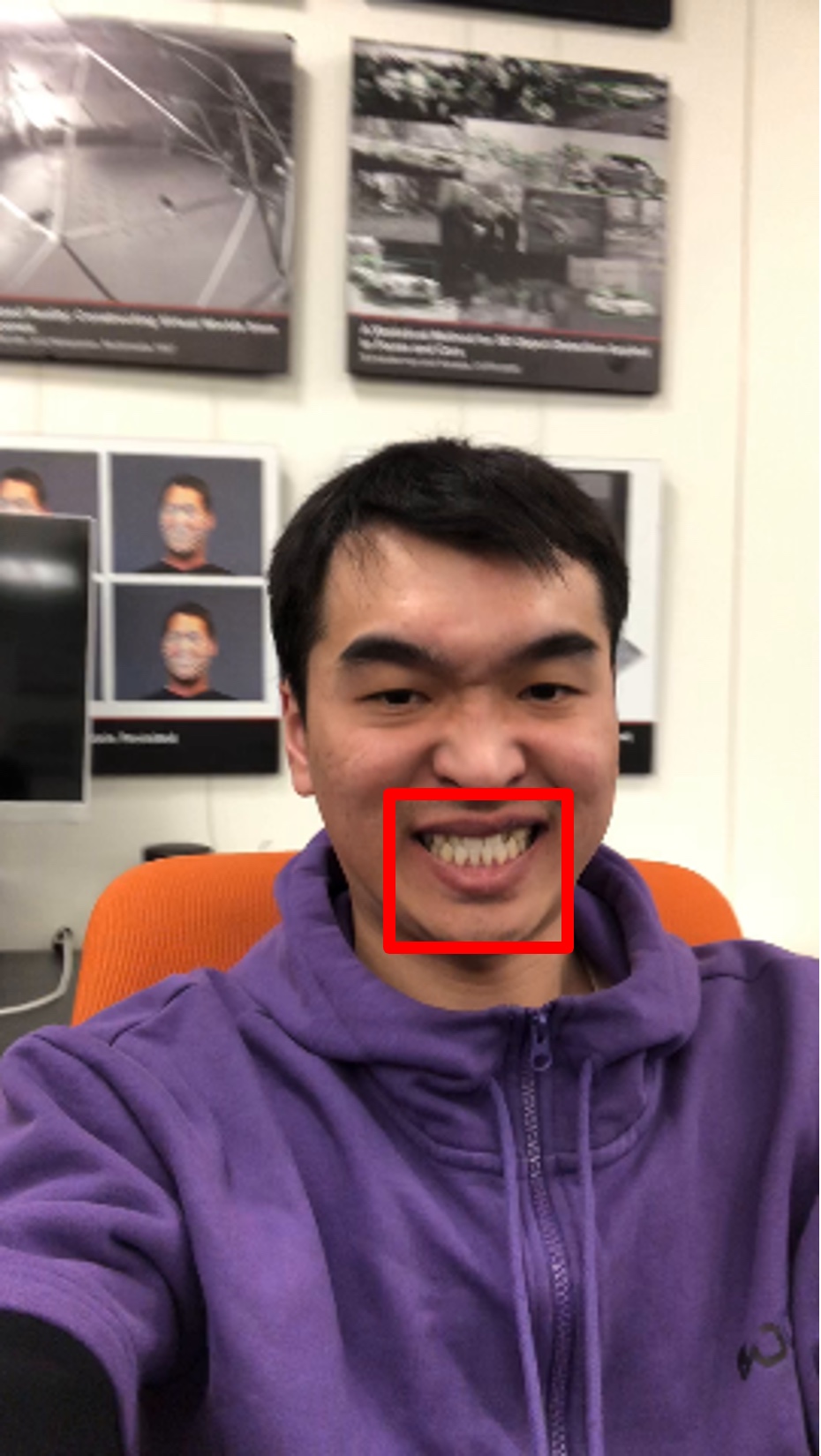}
         \caption*{\textbf{Expression}}
         \label{fig:exp}
     \end{subfigure}     
     \setcounter{subfigure}{0}
     \begin{subfigure}[b]{0.15\textwidth}
         \centering
         \includegraphics[width=0.99\textwidth]{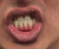}
         \caption{Ground Truth}
         \label{fig:exp-gt}
     \end{subfigure}
     \begin{subfigure}[b]{0.15\textwidth}
         \centering
         \includegraphics[width=0.99\textwidth]{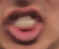}
         \caption{HyperNeRF \cite{park2021hypernerf}}
         \label{fig:exp-hyper}
     \end{subfigure}
    \begin{subfigure}[b]{0.15\textwidth}
         \centering
         \includegraphics[width=0.99\textwidth]{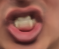}
         \caption{Ours-1}
         \label{fig:exp-ours1}
     \end{subfigure}
    \begin{subfigure}[b]{0.15\textwidth}
         \centering
         \includegraphics[width=0.99\textwidth]{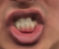}
         \caption{Ours-2}
         \label{fig:exp-ours2}
     \end{subfigure}
     \begin{subfigure}[b]{0.15\textwidth}
         \centering
         \includegraphics[width=0.99\textwidth]{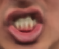}
         \caption{Ours-3}
         \label{fig:exp-ours3}
     \end{subfigure}
     \begin{subfigure}[b]{0.15\textwidth}
         \centering
         \includegraphics[width=0.99\textwidth,height=2.48cm,keepaspectratio]{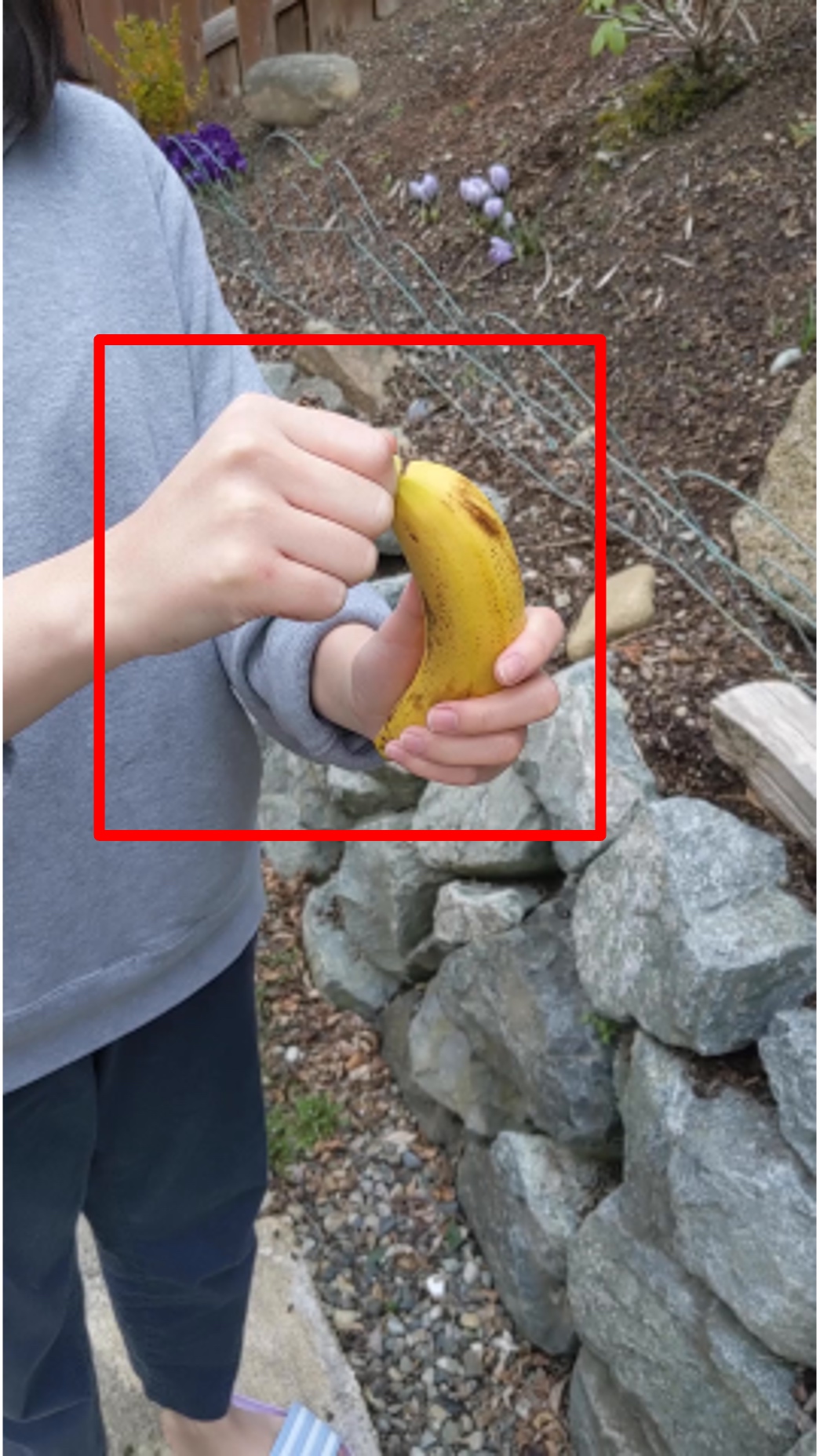}
         \caption*{\textbf{Peel Banana}}
         \label{fig:banana}
     \end{subfigure}     
     \begin{subfigure}[b]{0.15\textwidth}
         \centering
         \includegraphics[width=0.99\textwidth]{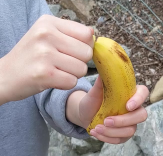}
         \caption{Ground Truth}
         \label{fig:banana-gt}
     \end{subfigure}
     \begin{subfigure}[b]{0.15\textwidth}
         \centering
         \includegraphics[width=0.99\textwidth]{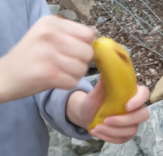}
         \caption{HyperNeRF \cite{park2021hypernerf}}
         \label{fig:banana-hyper}
     \end{subfigure}
    \begin{subfigure}[b]{0.15\textwidth}
         \centering
         \includegraphics[width=0.99\textwidth]{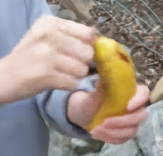}
         \caption{Ours-1}
         \label{fig:banana-ours1}
     \end{subfigure}
    \begin{subfigure}[b]{0.15\textwidth}
         \centering
         \includegraphics[width=0.99\textwidth]{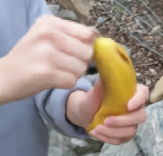}
         \caption{Ours-2}
         \label{fig:banana-ours2}
     \end{subfigure}
    \begin{subfigure}[b]{0.15\textwidth}
         \centering
         \includegraphics[width=0.99\textwidth]{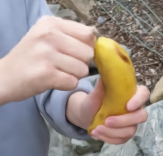}
         \caption{Ours-3}
         \label{fig:banana-ours3}
     \end{subfigure}

        \caption{Qualitative results for ablation on real dynamic scenes. We compare our DyLiN (Ours-1, Ours-2, Ours-3) with the ground truth and the HyperNeRF teacher model.
        Ours-1 was trained without our two MLPs.
        Ours-2 was trained with pointwise deformation MLP only.
        Ours-3 is our full model with both of our proposed two MLPs.}
        \label{fig:qual-real-ablation}
\end{figure*}

For the sake of completeness, \cref{fig:qual-real-ablation} illustrates qualitative ablation results for our model components on real dynamic scenes.
We found that sequentially adding our two proposed MLPs $T_\omega$ and $H_\psi$ improves the reconstruction, e.g., the gum between the teeth (\cref{fig:exp-ours3}) and the fingers (\cref{fig:banana-ours3}) become more and more apparent.
Without the MLPs, these parts were heavily blurred (\cref{fig:exp-ours1}, \cref{fig:banana-ours1}).

We kindly ask readers to refer to the supplementary material for CoDyLiN's qualitative results.

\section{Conclusion}
We proposed two architectures for extending LFNs to dynamic scenes.
Specifically, we introduced DyLiN, which models ray deformations without bending and lifts whole rays into a hyperspace, and CoDyLiN, which allows for controllable attribute inputs.
We trained both techniques via knowledge distillation from various dynamic NeRF teacher models.
We found that DyLiN produces state-of-the-art quality even without ray bending and CoDyLiN outperforms its teacher model,
while both are nearly 2 orders of magnitude faster than their strongest baselines.

Our methods do not come without limitations, however.
Most importantly, they focus on speeding up inference, as they require pretrained teacher models, which can be expensive to obtain.
In some experiments, our solutions were outperformed in terms of the PSNR score. Using the winners as teacher models could improve performance.
Additionally, distillation from multiple teacher models or joint training of the teacher and student models are also yet to be explored.
Moreover, we currently represent rays implicitly by sampling $K$ points along them, but increasing this number can lead to overfitting. An explicit ray representation may be more effective.
Finally, voxelizing and quantizing our models could improve efficiency.

Our results are encouraging steps towards achieving real-time volumetric rendering and animation, and we hope that our work will contribute to the progress in these areas.

\section*{Acknowledgements}
This research was supported partially by Fujitsu. We thank Chaoyang Wang from Carnegie Mellon University for the helpful discussion.

{\small
\bibliographystyle{ieee_fullname}
\bibliography{egbib}
}

\end{document}


\title{DyLiN: Making Light Field Networks Dynamic\\ Supplementary Material
}

\author{Heng Yu$^{1}$ \quad Joel Julin$^{1}$  \quad Zolt\'{a}n \'{A}. Milacski$^{1}$\quad Koichiro Niinuma$^{2}$ \quad L\'{a}szl\'{o} A. Jeni$^{1}$ \vspace{4pt}\\
	$^1$Robotics Institute, Carnegie Mellon University \quad
    $^2$Fujitsu Research of America \\
    {\tt\small \{hengyu, jjulin, zmilacsk\}@andrew.cmu.edu} \quad {\tt\small kniinuma@fujitsu.com} \quad {\tt\small laszlojeni@cmu.edu} \\
}
\maketitle

\section{Overview}
In this supplementary material, we provide detailed quantitative and additional qualitative results, showcasing the benefits of our proposed DyLiN and CoDyLiN methods. Furthermore, we also provide the training times one should expect given our current setup.

\section{Per-Scene Quantitative Results}
For the sake of completeness, we provide the detailed per-scene quantitative results for reconstruction quality (PSNR, SSIM, MS-SSIM, LPIPS) on the synthetic (\cref{tab:synth2}) and real (\cref{tab:real2}) dynamic scenes, extending Tab.~\textcolor{red}{1} and Tab.~\textcolor{red}{2} in the main paper that average these numbers across the scenes.
Accordingly, we found that our DyLiN performs the best with respect to the SSIM and LPIPS metrics, generating perceptually better images, yet it sometimes falls behind in terms of PSNR and MS-SSIM that may prefer blurred results.
Knowledge distillation improves a lot, our deformation and hyperspace MLPs yield slightly better results, while fine-tuning on the original training data gives a considerable boost.

\setcounter{table}{4}
\setcounter{figure}{8}

\begin{table*}[ht]
\centering
\caption{Per-scene quantitative results on synthetic dynamic scenes. Notations: Multi-Layer Perceptron (MLP), PD (pointwise deformation), FT (fine-tuning). We utilized D-NeRF as the teacher model for our DyLiNs. The winning numbers are highlighted in bold.}
\resizebox{\textwidth}{!}{%
\begin{tabular}{lcccccccccccc}
\toprule
                                 & \multicolumn{3}{c}{Hell Warrior}                              & \multicolumn{3}{c}{Mutant}                                     & \multicolumn{3}{c}{Hook}                                          & \multicolumn{3}{c}{Bouncing Balls}                                     \\
\cmidrule(lr){2-4} \cmidrule(lr){5-7} \cmidrule(lr){8-10} \cmidrule(lr){11-13}
Method                           & PSNR$\uparrow$           & SSIM$\uparrow$ & LPIPS$\downarrow$ & PSNR$\uparrow$            & SSIM$\uparrow$ & LPIPS$\downarrow$ & PSNR$\uparrow$               & SSIM$\uparrow$ & LPIPS$\downarrow$ & PSNR$\uparrow$                    & SSIM$\uparrow$ & LPIPS$\downarrow$ \\ \midrule
NeRF\cite{mildenhall2021nerf}    & 13.52                    & 0.8100           & 0.2500              & 20.31                     & 0.9100           & 0.0900              & 16.65                        & 0.8400           & 0.1900              & 20.26                             & 0.9100           & 0.2000              \\
DirectVoxGo\cite{SunSC22}        & 13.51                    & 0.7500           & 0.2500              & 19.45                     & 0.8900           & 0.1200              & 16.16                        & 0.8000           & 0.2100              & 20.20                             & 0.8700           & 0.2200              \\
Plenoxels\cite{fridovich2022plenoxels}& 15.19               & 0.7800           & 0.2700              & 21.44                     & 0.9100           & 0.0900              & 17.90                        & 0.8100           & 0.2100              & 21.30                             & 0.8900           & 0.1800              \\
T-NeRF\cite{pumarola2021d}       & 23.19                    & 0.9300           & 0.0800              & 30.56                     & 0.9600           & 0.0400              & 27.21                        & 0.9400           & 0.0600              & 37.81                             & 0.9800           & 0.1200              \\
D-NeRF\cite{pumarola2021d}       & 25.10                    & 0.9500           & 0.0600              & 31.29                     & 0.9700           & 0.0200              & 29.25                        & 0.9600           & 0.1100              & 38.93                             & 0.9800           & 0.1000              \\
TiNeuVox-S\cite{tineuvox}        & 27.00                    & 0.9500           & 0.0900              & 31.09                     & 0.9600           & 0.0500              & 29.30                        & 0.9500           & 0.0700              & 39.05                             & 0.9900           & 0.0600              \\
TiNeuVox-B\cite{tineuvox}        & \textbf{28.17}                   & 0.9700           & 0.0700              & 33.61                     & 0.9800           & 0.0300              & \textbf{31.45}                        & 0.9700           & 0.0500              & 40.73                             & 0.9900           & 0.0400              \\ \midrule
DyLiN, w/o two MLPs, w/o FT (ours)           &           26.81         &       0.9885         &       0.0363              &                     32.13      &         0.9961       &        0.0186          &                         
         29.89  &             0.9922      &          0.0297                         &    
        39.78 &       0.9997 &        0.0099                   \\
DyLiN, w/o two MLPs (ours)     &         27.73         &       0.9893         &       0.0317           &                     33.26  &        0.9971        &        0.0101           &              
        30.20        &    0.9928            &     0.0187              &            
        41.13 &       \textbf{0.9998} &      0.0064                 \\
DyLiN, PD MLP only, w/o FT (ours)           &          26.82      &           0.9886      &           0.0362        &                    32.13       &     0.9963           &      0.0185             &      
        29.94     &         0.9923       &     0.0296       &                 
        39.70 &           0.9996 &          0.0096              \\
DyLiN, PD MLP only (ours)     &            27.75      &      0.9896      &      0.0302           &                     33.47      &      0.9972          &       0.0102            &      
        30.39     &        0.9930         &         \textbf{0.0186}          &       
        41.52      &         \textbf{0.9998}      &         \textbf{0.0062}                \\
DyLiN, w/o FT (ours)      &             26.90        &       0.9887        &       0.0360            &                      32.17     &        0.9963        &        0.0182           &         
        29.99     &      0.9923          &         0.0289          &         
        40.02      &         0.9997      &         0.0098               \\
DyLiN (ours)  &               27.79      &        \textbf{0.9898}     &        \textbf{0.0298}        &                      \textbf{33.80}    &        \textbf{0.9974}        &         \textbf{0.0086}          &                       30.49       &      \textbf{0.9931}          &       \textbf{0.0186}            &    
        \textbf{41.59}       &     \textbf{0.9998}        &     \textbf{0.0062}                \\\midrule
                                 & \multicolumn{3}{c}{Lego} &                                    \multicolumn{3}{c}{T-Rex} &                                    \multicolumn{3}{c}{Stand Up} &                                   \multicolumn{3}{c}{Jumping Jacks}                                    \\
\cmidrule(lr){2-4} \cmidrule(lr){5-7} \cmidrule(lr){8-10} \cmidrule(lr){11-13}
Method                           & PSNR$\uparrow$           & SSIM$\uparrow$ & LPIPS$\downarrow$ & PSNR$\uparrow$            & SSIM$\uparrow$ & LPIPS$\downarrow$ & PSNR$\uparrow$               & SSIM$\uparrow$ & LPIPS$\downarrow$ & PSNR$\uparrow$                    & SSIM$\uparrow$ & LPIPS$\downarrow$ \\ \midrule
NeRF\cite{mildenhall2021nerf}    & 20.30                    & 0.7900           & 0.2300              & 24.29                     & 0.9300           & 0.1300              & 18.19                        & 0.8900           & 0.1400              & 18.28                             & 0.8800           & 0.2300              \\
DirectVoxGo\cite{SunSC22}        & 21.13                    & 0.9000           & 0.1000              & 23.27                     & 0.9200           & 0.0900              & 17.58                        & 0.8600           & 0.1600              & 17.80                             & 0.8400           & 0.2000              \\
Plenoxels\cite{fridovich2022plenoxels}& 21.97                    & 0.9000           & 0.1100              & 25.18                     & 0.9300           & 0.0800              & 18.76                        & 0.8700           & 0.1500              & 20.18                             & 0.8600           & 0.1900              \\
T-NeRF\cite{pumarola2021d}       & 23.82                    & 0.9000           & 0.1500              & 30.19                     & 0.9600           & 0.1300              & 31.24                        & 0.9700           & 0.0200              & 32.01                             & 0.9700          & 0.0300              \\
D-NeRF\cite{pumarola2021d}       & 21.64                    & 0.8300           & 0.1600              & 31.75                     & 0.9700           & 0.0300              & 32.79                        & 0.9800           & 0.0200              & 32.80                             & 0.9800           & 0.0300              \\
TiNeuVox-S\cite{tineuvox}        & 24.35                    & 0.8800           & 0.1300              & 29.95                     & 0.9600           & 0.0600              & 32.89                        & 0.9800           & 0.0300              & 32.33                             & 0.9700           & 0.0400              \\
TiNeuVox-B\cite{tineuvox}        & \textbf{25.02}                   & 0.9200           & 0.0700              & 32.70                     & 0.9800           & 0.0300              & 35.43                        & 0.9900           & 0.0200              & \textbf{34.23}                             & 0.9800           & 0.0300              \\ \midrule
DyLiN, w/o two MLPs, w/o FT (ours)         &    22.11                &        0.9747        &   0.0612                &     
31.35    &      0.9978   &       0.0290              &            
33.98     &        0.9973     &        0.0140            &                   33.24 & 0.9981 & 0.0260              \\
DyLiN, w/o two MLPs (ours)    &    22.42                 &       0.9761        &   0.0493           &        
32.80      &      0.9984      &      0.0170               &               35.31     &      0.9980     &      0.0084            &                 33.67   &  0.9984   &  0.0155                   \\
DyLiN, PD MLP only, w/o FT (ours)          &    22.13               &       0.9748         &      0.0618          &          
32.18     &      0.9982     &      0.0282          &                    33.97    &      0.9973    &      0.0140            &                 33.19 &  0.9982 &  0.0257                \\
DyLiN, PD MLP only (ours)    &    22.76                &       0.9775     &        0.0452       &        
32.77     &          \textbf{0.9985}     &          0.0176              &          35.56      &  0.9981      &    0.0082               &                    33.68 &  0.9984 &  0.0152                 \\
DyLiN, w/o FT (ours)      &     22.24              &       0.9754        &        0.0600      &           
32.24     &      0.9982     &      0.0276           &                    34.15    &     0.9974    &     0.0141             &                              33.23  &   0.9983  &   0.0256              \\
DyLiN (ours)  &       23.10         &     \textbf{0.9791}   & \textbf{0.0443}  &          
\textbf{32.91}  &  \textbf{0.9985}   &  \textbf{0.0168}    &  \textbf{35.95}   &    \textbf{0.9983}   &    \textbf{0.0074}         &                                  33.84 &  \textbf{0.9985} &  \textbf{0.0151}   \\    
\bottomrule
\end{tabular}%
}
\label{tab:synth2}
\end{table*}


\begin{table*}[ht]
\centering
\caption{Per-scene quantitative results on real dynamic scenes. Notations: Multi-Layer Perceptron (MLP), PD (pointwise deformation), FT (fine-tuning), N/A (not available in the cited research paper). We utilized HyperNeRF as the teacher model for our DyLiNs. The winning numbers are highlighted in bold.}
\resizebox{\textwidth}{!}{%
\begin{tabular}{lcccccc}
\toprule
                                                 & \multicolumn{2}{c}{Broom}          & \multicolumn{2}{c}{3D Printer}     & \multicolumn{2}{c}{Chicken}        \\
\cmidrule(lr){2-3} \cmidrule(lr){4-5} \cmidrule(lr){6-7}                    Method                                           & PSNR$\uparrow$ & MS-SSIM$\uparrow$ & PSNR$\uparrow$ & MS-SSIM$\uparrow$ & PSNR$\uparrow$ & MS-SSIM$\uparrow$ \\ \midrule
NeRF\cite{mildenhall2021nerf}   & 19.90           & 0.653             & 20.70           & 0.780             & 19.90           & 0.777             \\
NV    \cite{Lombardi:2019}                                             & 17.70           & 0.623             & 16.20           & 0.665             & 17.60           & 0.615             \\
NSFF     \cite{li2020neural}                                        & \textbf{26.10}          & \textbf{0.871}             & \textbf{27.70}           &\textbf{0.947}             & 26.90           & 0.944             \\
Nerfies     \cite{park2021nerfies}                                        & 19.20           & 0.567             & 20.60           & 0.830             & 26.70           & 0.943             \\
HyperNeRF   \cite{park2021hypernerf}                                          & 19.30           & 0.591             & 20.00           & 0.821             & 26.90           & 0.948             \\
TiNeuVox-S\cite{tineuvox}       & 21.90           & 0.707             & 22.70           & 0.836             & 27.00           & 0.929             \\
TiNeuVox-B\cite{tineuvox}       & 21.50           & 0.686             & 22.80           & 0.841             & \textbf{28.30}           & 0.947             \\ \midrule
DyLiN, w/o two MLPs, w/o FT (ours)                            &           21.98 & 0.808            &          22.99 & 0.899            &  26.89    &    0.948               \\
DyLiN, w/o two MLPs (ours)                    &        22.04  &  0.811         &            23.16 & 0.905           &  27.35    &   0.954            \\
DyLiN, PD MLP only, w/o FT (ours)                          &       22.02 & 0.805       &                23.04 & 0.903          &  26.88    & 0.948                  \\
DyLiN, PD MLP only (ours)                      &      22.14 & 0.815           &              23.19 & 0.906             &  27.53     & 0.955              \\
DyLiN, w/o FT (ours)                       &     22.04 & 0.809            &               23.06 & 0.902         &  26.91   &   0.948              \\
DyLiN (ours)                &      22.14 & 0.823            &              23.21 & 0.906          &  27.62   &  \textbf{0.956}       \\ \midrule
                                                 & \multicolumn{2}{c}{Peel Banana}    & \multicolumn{2}{c}{Americano}      & \multicolumn{2}{c}{Expressions}    \\
\cmidrule(lr){2-3} \cmidrule(lr){4-5} \cmidrule(lr){6-7}
Method                                           & PSNR$\uparrow$ & MS-SSIM$\uparrow$ & PSNR$\uparrow$ & MS-SSIM$\uparrow$ & PSNR$\uparrow$ & MS-SSIM$\uparrow$ \\ \midrule
NeRF\cite{mildenhall2021nerf}   & 20.00           & 0.769             &       N/A         &       N/A          &          N/A    &         N/A        \\
NV \cite{Lombardi:2019}                                                  & 15.90           & 0.380             &      N/A        &        N/A         &          N/A    &         N/A        \\
NSFF  \cite{li2020neural}                                           & 24.60           & 0.902             &      N/A        &       N/A          &          N/A    &         N/A        \\
Nerfies   \cite{park2021nerfies}                                         & 22.40           & 0.872             &      N/A        &       N/A          &          N/A    &         N/A        \\
HyperNeRF   \cite{park2021hypernerf}                                     & 23.30           & 0.896             &     18.42           &        0.720           &         25.40       &   0.958                \\
TiNeuVox-S\cite{tineuvox}       & 22.10           & 0.780             &      N/A        &       N/A          &           N/A   &         N/A        \\
TiNeuVox-B\cite{tineuvox}       & 24.40           & 0.873             &      N/A        &       N/A          &           N/A   &         N/A        \\ \midrule
DyLiN, w/o two MLPs, w/o FT (ours)    &       23.38 & 0.872        &   18.45     &  0.722                  &       25.36 & 0.950    \\
DyLiN, w/o two MLPs (ours)                      &         24.35 & 0.906        &          30.85    &           0.977     &           26.33 & 0.967            \\
DyLiN, PD MLP only, w/o FT (ours)                            &          23.70 & 0.882       &         18.47 &  0.722            &            25.55 & 0.960          \\
DyLiN, PD MLP only (ours)                     &       25.72 & 0.936           &        31.01 & 0.978        &           26.33 & 0.967           \\
DyLiN, w/o FT (ours)                       &        23.97 & 0.886            &         18.48 & 0.722            &           26.51 & 0.969         \\
DyLiN (ours)                &         \textbf{27.36} & \textbf{0.952}           &          \textbf{31.56} & \textbf{0.982}         &        \textbf{26.91} & \textbf{0.974}  \\
\bottomrule
\end{tabular}%
}
\label{tab:real2}
\end{table*}

\section{More Qualitative Results}
We provide additional qualitative results for 3 experiments.

\begin{figure*}[!ht]
     \centering
     \begin{subfigure}[b]{0.15\textwidth}
         \centering
         \includegraphics[width=0.99\textwidth]{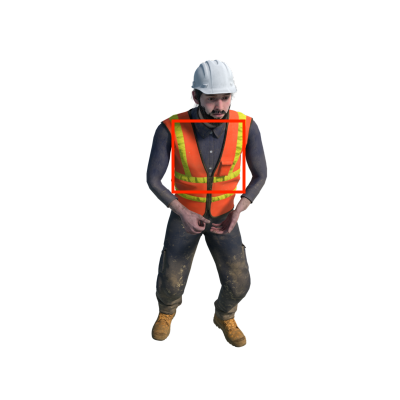}
         \caption*{\textbf{Standup}}
         \label{fig:hook_box}
     \end{subfigure}     
     \setcounter{subfigure}{0}
     \begin{subfigure}[b]{0.15\textwidth}
         \centering
         \includegraphics[width=0.99\textwidth]{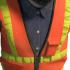}
         \caption{Ground Truth}
         \label{fig:hook_gt}
     \end{subfigure}
     \begin{subfigure}[b]{0.15\textwidth}
         \centering
         \includegraphics[width=0.99\textwidth]{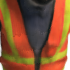}
         \caption{D-NeRF \cite{pumarola2021d}}
         \label{fig:hook_dnerf}
     \end{subfigure}
    \begin{subfigure}[b]{0.15\textwidth}
         \centering
         \includegraphics[width=0.99\textwidth]{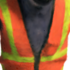}
         \caption{TiNeuVox \cite{tineuvox}}
         \label{fig:hook_tineuvox}
     \end{subfigure}
    \begin{subfigure}[b]{0.15\textwidth}
         \centering
         \includegraphics[width=0.99\textwidth]{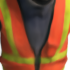}
         \caption{Ours-1}
         \label{fig:hook_ours1}
     \end{subfigure}
    \begin{subfigure}[b]{0.15\textwidth}
         \centering
         \includegraphics[width=0.99\textwidth]{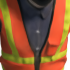}
         \caption{Ours-2}
         \label{fig:hook_ours2}
     \end{subfigure}    
     
     \begin{subfigure}[b]{0.15\textwidth}
         \centering
         \includegraphics[width=0.99\textwidth]{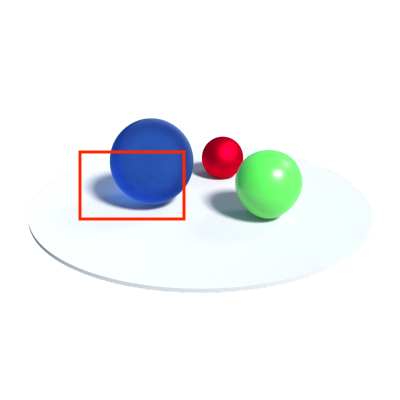}
         \caption*{\textbf{Bouncing Ball}}
         \label{fig:jumping_box}
     \end{subfigure}     
     \begin{subfigure}[b]{0.15\textwidth}
         \centering
         \includegraphics[width=0.99\textwidth]{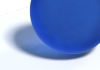}
         \caption{Ground Truth}
         \label{fig:jumping_gt}
     \end{subfigure}
     \begin{subfigure}[b]{0.15\textwidth}
         \centering
         \includegraphics[width=0.99\textwidth]{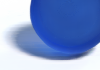}
         \caption{D-NeRF \cite{pumarola2021d}}
         \label{fig:jumping_dnerf}
     \end{subfigure}
    \begin{subfigure}[b]{0.15\textwidth}
         \centering
         \includegraphics[width=0.99\textwidth]{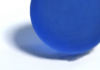}
         \caption{TiNeuVox \cite{tineuvox}}
         \label{fig:jumping_tineuvox}
     \end{subfigure}
    \begin{subfigure}[b]{0.15\textwidth}
         \centering
         \includegraphics[width=0.99\textwidth]{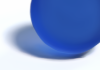}
         \caption{Ours-1}
         \label{fig:jumping_ours1}
     \end{subfigure}
    \begin{subfigure}[b]{0.15\textwidth}
         \centering
         \includegraphics[width=0.99\textwidth]{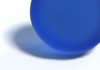}
         \caption{Ours-2}
         \label{fig:jumping_ours2}
     \end{subfigure}

        \caption{More qualitative results on synthetic dynamic scenes. We compare our DyLiN (Ours-1, Ours-2) with the ground truth, the D-NeRF teacher model and TiNeuVox.
        Ours-1 and Ours-2 were trained without and with fine-tuning on the original data, respectively.}
        \label{fig:qual_synth}
\vspace{\intextsep}
     \begin{subfigure}[b]{0.15\textwidth}
         \centering
         \includegraphics[width=0.99\textwidth,height=2.492cm,keepaspectratio]{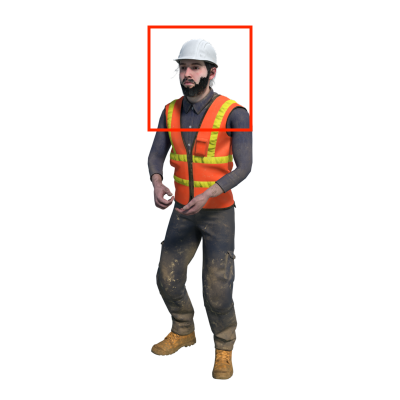}
         \caption*{\textbf{Standup}}
         \label{fig:exp2}
     \end{subfigure}     
     \setcounter{subfigure}{0}
     \begin{subfigure}[b]{0.15\textwidth}
         \centering
         \includegraphics[width=0.99\textwidth]{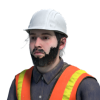}
         \caption{Ground Truth}
         \label{fig:exp2-gt}
     \end{subfigure}
     \begin{subfigure}[b]{0.15\textwidth}
         \centering
         \includegraphics[width=0.99\textwidth]{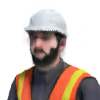}
         \caption{D-NeRF \cite{pumarola2021d}}
         \label{fig:exp2-hyper}
     \end{subfigure}
    \begin{subfigure}[b]{0.15\textwidth}
         \centering
         \includegraphics[width=0.99\textwidth]{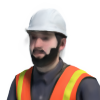}
         \caption{Ours-1}
         \label{fig:exp2-ours1}
     \end{subfigure}
    \begin{subfigure}[b]{0.15\textwidth}
         \centering
         \includegraphics[width=0.99\textwidth]{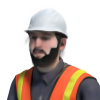}
         \caption{Ours-2}
         \label{fig:exp2-ours2}
     \end{subfigure}
     \begin{subfigure}[b]{0.15\textwidth}
         \centering
         \includegraphics[width=0.99\textwidth]{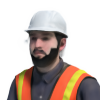}
         \caption{Ours-3}
         \label{fig:exp2-ours3}
     \end{subfigure}     

        \caption{Qualitative results for ablation on the synthetic Standup scene. We compare our DyLiN (Ours-1, Ours-2, Ours-3) with the ground truth and the D-NeRF teacher model.
        Ours-1 was trained without our two MLPs.
        Ours-2 was trained with pointwise deformation MLP only.
        Ours-3 is our full model with both of our proposed two MLPs.}
        \label{fig:qual-real-ablation}
\end{figure*}

First, \cref{fig:qual_synth} depicts more qualitative results for reconstruction quality on synthetic dynamic scenes, extending Fig.~\textcolor{red}{6} in the main paper.
Specifically, the Standup scene includes buttons on the shirt of the avatar (\cref{fig:hook_gt}), and the baselines are all missing them (\cref{fig:hook_dnerf,fig:hook_tineuvox}), whereas our full method is capable of reconstructing such details (\cref{fig:hook_ours2}).
Furthermore, the Bouncing Ball scene involves shadow casting (\cref{fig:jumping_gt}).
Inside the shadowed area, D-NeRF \cite{pumarola2021d} produces horizontal artifacts (\cref{fig:jumping_dnerf}), while TiNeuVox \cite{tineuvox} predicts an inaccurate boundary (\cref{fig:jumping_tineuvox}).
Again, our full model outputs the correct shadow (\cref{fig:jumping_ours2}).

Second, \cref{fig:qual-real-ablation} shows qualitative results for ablation on the synthetic Standup scene using a D-NeRF teacher model, complementing Fig.~\textcolor{red}{8} in the main paper that is restricted to real scenes and distilling from HyperNeRF \cite{park2021hypernerf}.
D-NeRF gives an oversmoothed prediction (\cref{fig:exp2-hyper}), whereas the two MLPs of our DyLiN gradually reduce the blurriness of the face (\cref{fig:exp2-ours1,fig:exp2-ours2,fig:exp2-ours3}).

\begin{figure*}[!htb]
\centering
 \includegraphics[width=0.99\linewidth]{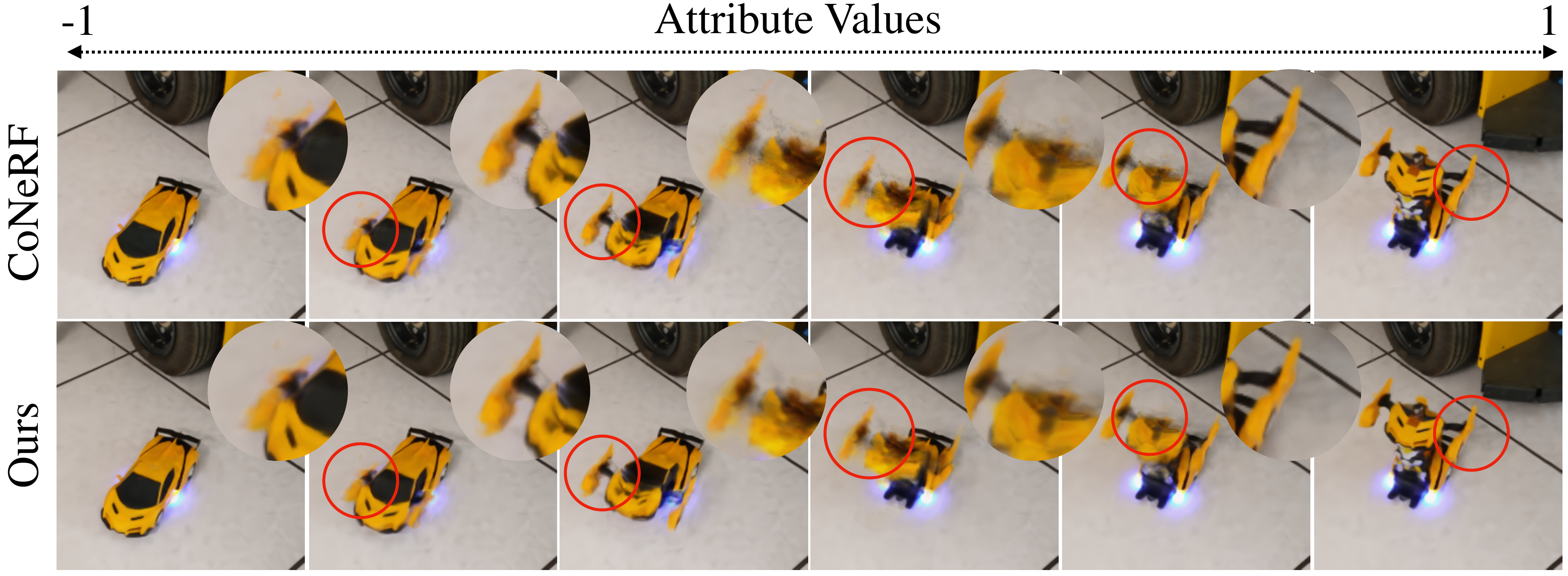}
  \captionof{figure}{Qualitative results on the real controllable Transformer scene.
  We utilized CoNeRF \cite{kania2022conerf} as the teacher model for our CoDyLiN. 
  Red circles indicate regions enlarged in insets.
  Best viewed zoomed in.}
 \label{fig:cor2l_res}
\end{figure*}

Lastly, \cref{fig:cor2l_res} illustrates qualitative results for the real controllable Transformer scene, complementing the numbers of Tab.~\textcolor{red}{4} in the main paper.
We portray the effects of altering the attribute input $\alpha_i\in[-1,1]$, which encodes the body pose of the character.
We found that the CoNeRF \cite{kania2022conerf} teacher model produces yellow color artifacts outside the boundary of the character (see, e.g., top row 1\textsuperscript{st} inset), whereas our CoDyLiN student model captures the boundary well.
\section{Training Times}
On a single NVIDIA A100 GPU, the full process takes $\approx 38\text{--}\SI{43}{\hour}$, including $5\text{--}\SI{10}{\hour}$ to train the teacher, $\SI{13}{\hour}$ for drawing $S=\SI{10000}{}$ training samples for KD, and $\SI{20}{\hour}$ for training the student via KD.

\bibliographystyle{ieee_fullname}
\bibliography{egbib}